\documentclass[lettersize,journal]{IEEEtran}
\usepackage[T1]{fontenc}
\usepackage{graphicx}
\usepackage{subcaption}
\usepackage{color}
\usepackage{textcomp}
\usepackage{csquotes}
\usepackage{hyperref}
\usepackage[american]{circuitikz} 

\usepackage{caption} 
\usepackage{tabularx}
\usepackage{float}
\usepackage{array}  
\usepackage{booktabs} 
\usepackage{multirow}
\usepackage{makecell}
\usepackage{arydshln} 
\usepackage{todonotes}
\usepackage{eurosym}

\usepackage{siunitx}
\usepackage{cite}
\begin{document}



\title{Touch Speaks, Sound Feels: A Multimodal Approach to Affective and Social Touch from Robots to Humans.}


\author{Qiaoqiao Ren \and and
Tony Belpaeme

\thanks{Faculty of Engineering and Architecture, IDLab-AIRO, Ghent University -- imec, Technologiepark 126, 9052 Gent, Belgium}

\thanks{The authors acknowledge the use of generative AI in preparing this manuscript. Specifically, Grammarly and GPT-4o were used to assist with grammar checking and enhancing the overall readability. All content was subsequently reviewed and edited by the authors, who accept full responsibility for the final version of the manuscript.}



}
\maketitle

\begin{abstract}

Affective tactile interaction constitutes a fundamental component of human communication. In natural human–human encounters, touch is seldom experienced in isolation; rather, it is inherently multisensory. Individuals not only perceive the physical sensation of touch but also register the accompanying auditory cues generated through contact. The integration of haptic and auditory information forms a rich and nuanced channel for emotional expression. While extensive research has examined how robots convey emotions through facial expressions and speech, their capacity to communicate social gestures and emotions via touch remains largely underexplored. To address this gap, we developed a multimodal interaction system incorporating a 5×5 grid of 25 vibration motors synchronized with audio playback of touch sound, enabling robots to deliver combined haptic-audio stimuli. In an experiment involving 32 Chinese participants, ten emotions and six social gestures were presented through vibration, sound, or their combination. Participants rated each stimulus on arousal and valence scales. The results revealed that (1) the combined haptic-audio modality significantly enhanced decoding accuracy compared to single modalities; (2) each individual channel—vibration or sound—effectively supported certain emotions recognition, with distinct advantages depending on the emotional expression; and (3) gestures alone were generally insufficient for conveying clearly distinguishable emotions. These findings underscore the importance of multisensory integration in affective human–robot interaction and highlight the complementary roles of haptic and auditory cues in enhancing emotional communication.

\end{abstract}


\section{Introduction}

\IEEEPARstart{T}{o}uch is a fundamental aspect of human interaction, serving as a primary channel for communication, emotional expression, and social bonding \cite{field2014touch}. Unlike other sensory modalities such as vision or hearing, touch provides direct and immediate feedback, reinforcing trust, intimacy, and empathy in interpersonal relationships \cite{della2022interpersonal}. Studies in psychology and neuroscience have shown that affective touch, such as a gentle pat or a reassuring squeeze, can reduce stress, enhance well-being, and strengthen social bonds \cite{banissy2023touch, field2014touch}. Early research on human-human affective tactile communication suggests that touch primarily conveys emotional valence and intensity, with context playing a critical role in interpretation \cite{eid2015affective}. Hertenstein \textit{et al.} demonstrated that emotions such as anger, disgust, fear, gratitude, happiness, love, sadness, and sympathy could be recognized solely through touch, with recognition rates between 48\% and 83\%, whereas self-focused emotions (e.g., embarrassment, envy, pride) were less accurately conveyed \cite{hertenstein2006touch}. Studies have shown that haptic behaviour alone is insufficient for distinct emotion recognition, with research identifying 23 different types of tactile behaviours (e.g., hugging, squeezing, shaking) that do not exclusively correspond to a single emotion. For instance, stroking was observed in expressions of sadness, love, and sympathy, suggesting that tactile behaviours must be interpreted alongside other features such as duration and intensity to enhance emotion decoding. \cite{hertenstein2009communication}.

As robots become increasingly integrated into daily life—from caregiving and therapy to collaborative workspaces, establishing effective communication between humans and robots is essential \cite{vercelli2018robots}. While current research in human-robot affective interaction has predominantly emphasized modalities such as speech \cite{moriyama1999emotion}, facial expressions \cite{tarnowski2017emotion}, and body movement \cite{ahmed2019emotion}, the tactile channel remains relatively underexplored \cite{melaisi2018multimodal}. This neglect is partly due to the interdisciplinary challenges inherent in studying affective touch \cite{eid2015affective}. However, prior studies have shown that tactile interaction with robots can significantly influence human behaviours, such as risk-taking, stress responses, and attitudes toward the robot \cite{ren2024tactile}. 


Recent research has increasingly explored the role of vibration and sound in emotion expression, demonstrating that vibrotactile signals can effectively convey emotions such as happiness, sadness, and anger \cite{ju2021haptic}. Specific vibration patterns, defined by variations in frequency, amplitude, and rhythm, have been mapped to distinct emotional states, contributing to affective perception \cite{seifi2018toward}. Notably, touch is rarely experienced in isolation; it is inherently multisensory \cite{fulkerson2014rethinking}. When interacting with vibrotactile systems, individuals often experience not only the tactile sensations but also the accompanying auditory cues produced by touch. Such as snatching or tapping. Recent research, for instance, has investigated how people decode emotional intent and specific touch gestures from these tactile sounds. This opens new avenues for capturing the quality of social touch \cite{de2025paving}.

This co-occurrence of tactile and auditory feedback has important implications for affective human-robot interaction (HRI). While computer vision has been widely adopted for emotion recognition, alternative modalities become essential in contexts where visual input is limited or unreliable \cite{amirova202110}. Auditory signals, based on their temporal and spectral properties, can independently convey affective qualities such as urgency, warmth, or discomfort \cite{tajadura2008embodied}. Moreover, previous studies further indicate that tactile gestures and emotional expressions can be recognized through their auditory counterparts alone \cite{de2023audio}. Despite these insights, the potential of integrating tactile and auditory modalities during human-robot tactile interaction remains underexplored. Most existing studies focus solely on vision and sound \cite{ren2024no} for emotion expression \cite{rawal2022facial}, overlooking the communicative richness of multimodal tactile–auditory signals. There is limited understanding of how humans perceive and decode both the combination of touch and contact-generated sound, as well as each modality's contribution in affective and social contexts in HRI. Addressing this gap is crucial for developing emotionally intelligent robots capable of effective communication when visual cues are limited, and enabling remote interactions with both human partners and virtual agents.

Another key challenge in HRI is the physical limitation of current humanoid platforms. For example, robots such as Pepper and Nao offer only limited tactile expressivity, as their grippers are typically restricted to basic open–close motions. In contrast, advances in mediated touch technologies—such as haptic gloves, vibration devices, and smart bands—have demonstrated the effectiveness of tactile feedback in virtual and augmented reality, highlighting the potential for richer multisensory integration in HRI \cite{emami2024survey, kuhail2024advances}. However, these devices are rarely designed for emotional communication in HRI, and their integration with other sensory modalities remains limited. Furthermore, the perception and interpretation of affective touch can be influenced by the embodiment of the humanoid robot, making it valuable to investigate how such technologies might be adapted for human–robot interaction. To address this, we employ mediated tactile technologies, specifically, wearable vibrotactile devices that enable spatially distributed and dynamic feedback. In this context, we developed the VibroSleeve, a wearable haptic interface that delivers spatially distributed vibration to convey different emotions. By integrating vibrotactile and auditory modalities, this study proposes a multimodal framework for conveying affective and touch gestures from a humanoid robot to humans and investigates the following research questions:

\begin{enumerate}

    \item Does the combination of haptic and auditory cues generated by touch enhance the accuracy of emotion and touch gesture recognition compared to either modality alone?

    \item Do participants associate certain touch gestures with specific emotional meanings, even when those gestures were not explicitly designed to convey particular affective responses?

    \item What are the individual contributions of the haptic and auditory channels to emotion and touch gestures decoding, and how does their effectiveness vary across different emotional expressions?

    

    \item Which emotions or touch gestures are most frequently misinterpreted, and what patterns of confusion emerge across modalities?




\end{enumerate}

\section{Relevant Work}

Affective communication via mediated touch has gained increasing attention as a means of transmitting emotional information through haptic interfaces, either between humans or between humans and machines \cite{olugbade2023touch, ren2025touched}. A central question in this domain concerns how effectively users can recognize and interpret robotic emotions conveyed through mediated touch, particularly in comparison to human–human tactile interaction \cite{van2015social}.

Research into tactile interfaces spans a range of technologies, including wearable systems, robotic actuators, and mid-air haptic devices. For instance, Yohanan and MacLean developed the Haptic Creature, a social robot that conveys emotional states through multimodal tactile behaviours such as ear stiffness, breathing, and purring \cite{yohanan2011design}. In parallel, although force-feedback-based emotion communication remains relatively underexplored, foundational work by Smith and MacLean demonstrated that affective states can be transmitted solely through haptic means \cite{smith2007communicating}. Similarly, Bailenson \textit{et al.} showed that emotions such as anger, disgust, joy, fear, sadness, and surprise could be communicated using a haptic device, though with modest accuracy, notably lower than that achieved by people communicating emotions through handshakes \cite{bailenson2007virtual}. Bonnet \textit{et al.} extended these findings by integrating haptic feedback with visual avatars, which improved recognition rates for emotions like anger and disgust \cite{bonnet2011improvement}.

Empirical studies have shed light on the physical characteristics of affective touch. Huisman \textit{et al.} used a pressure-sensitive wearable sleeve to analyze how users express emotions through touch. They found that fear, happiness, and anger involved larger contact areas, sympathy was associated with greater force, and temporal patterns varied by emotion—e.g., anger had shorter intervals between touches than happiness or love \cite{huisman2013towards}. However, the study focused on expression rather than recognition of emotions.

A more targeted subset of wearable haptic devices has emerged to support interpersonal emotional communication. Examples include ``hug shirts'' that use distributed air actuators to simulate affective touch remotely \cite{cheok2019huggy}. Ju \textit{et al.} explored this domain by mapping touch-generated sounds to vibration amplitude, asking participants to express four emotions through the same gestures. Their system achieved a recognition rate of 57.74\% against a 25\% chance level \cite{ju2021haptic}. Bailenson’s earlier work examined mediated touch between humans using a 2-DOF force-feedback joystick to express seven emotions, achieving 33.04\% recognition accuracy compared to a 14.29\% chance level \cite{bailenson2007virtual}.

These studies collectively suggest that vibration, when carefully modulated in terms of frequency, amplitude, spatial location, and rhythm, offers a powerful channel for emotional expression. Dynamic vibrotactile signals enable fine-grained affective communication, making them highly relevant for social human–robot interaction. Moreover, combining vibrotactile feedback with auditory or visual cues can further enhance emotional clarity and social engagement.

\section{Materials \& Methods}

\begin{figure}
    \centering
        \includegraphics [width=\columnwidth]{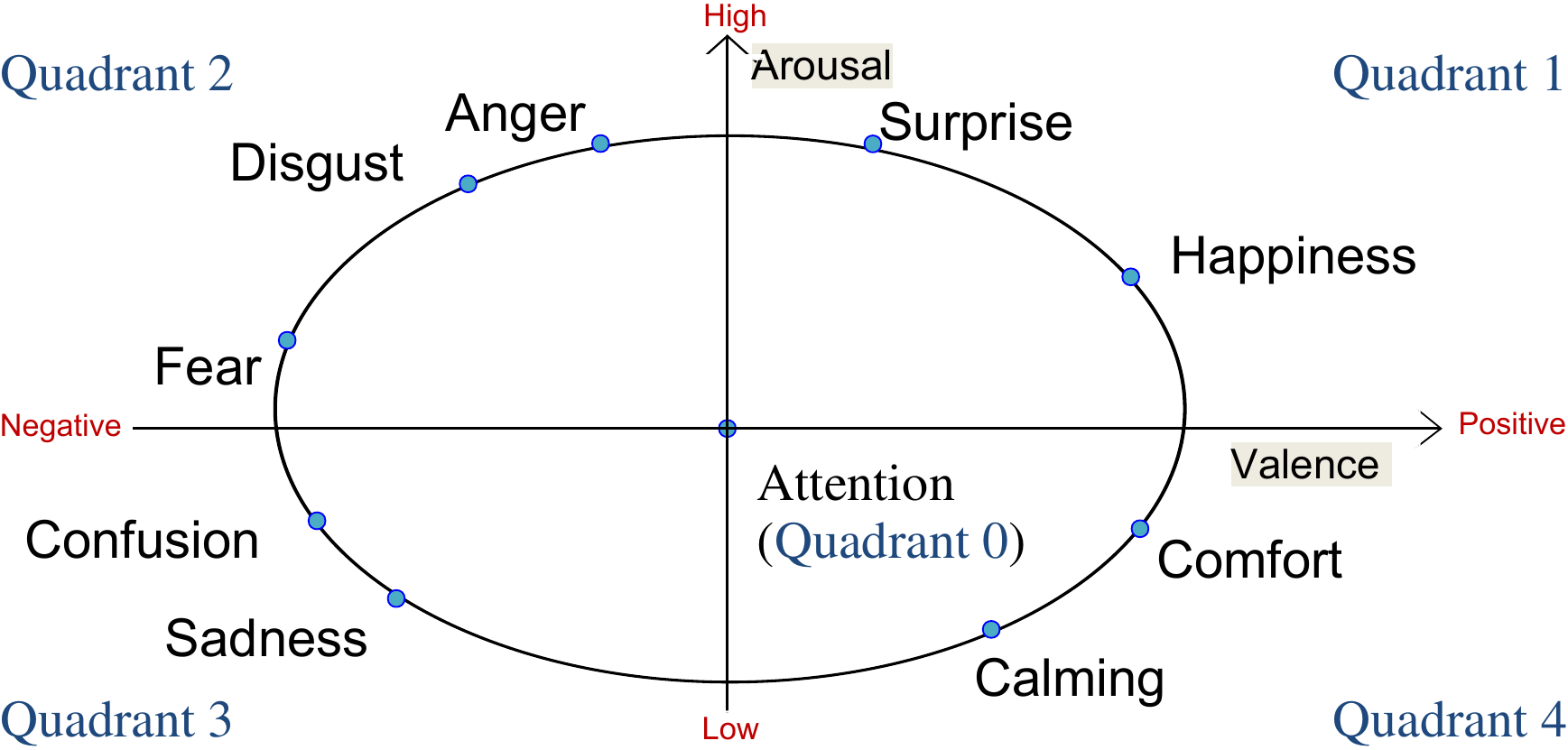}
    \caption{Emotions distribution based on Russell’s circumplex model. Emotions are positioned on Russell’s circumplex model according to their valence and arousal levels. The intention ``grab attention'' is not an emotion and is therefore placed at the origin (Quadrant 0), representing its status as a communicative act outside the affective dimensions.}
    \label{russel_model}
\end{figure}

In this experiment, we propose to collect data on how people interpret tactile stimuli. Participants will be tasked with decoding 10 emotions (anger, fear, disgust, happiness, surprise, sadness, confusion, comfort, calm, and attention) and 6 gestures (hold, pat, tickle, rub, tap, and poke). Before data collection, participants are provided with definitions of all emotions (to ensure consistent interpretations for non-native English-speaking participants) and touch gestures to ensure they understood and agreed with the given definitions. Participants are asked to rate the arousal, valence, and dominance for each stimulus drawn from our previous study \cite{ren2024conveying}. Participants receive definitions of arousal and valence, along with a detailed explanation of each session's procedure. The experiment begins only after they fully understand all terms and steps, with the option to ask for further clarification at any time. Afterwards, they are asked to choose a specific emotion or touch gesture for the stimulus. The following sections will detail the equipment used, the setup process, and the specific configurations implemented for data acquisition. According to previous research, we classified the 10 emotions into different arousal and valence quadrants based on Russell's circumplex model as shown in the \ref{russel_model}. In addition, some researchers state that surprise is valence-free, while other researchers argue that surprise should be either positive or negative, rather than being described as neither positive nor negative. Here, we put surprise emotions in the negative valence zones.

\begin{figure*}[h] 
    \centering 
    \includegraphics [width=\textwidth]{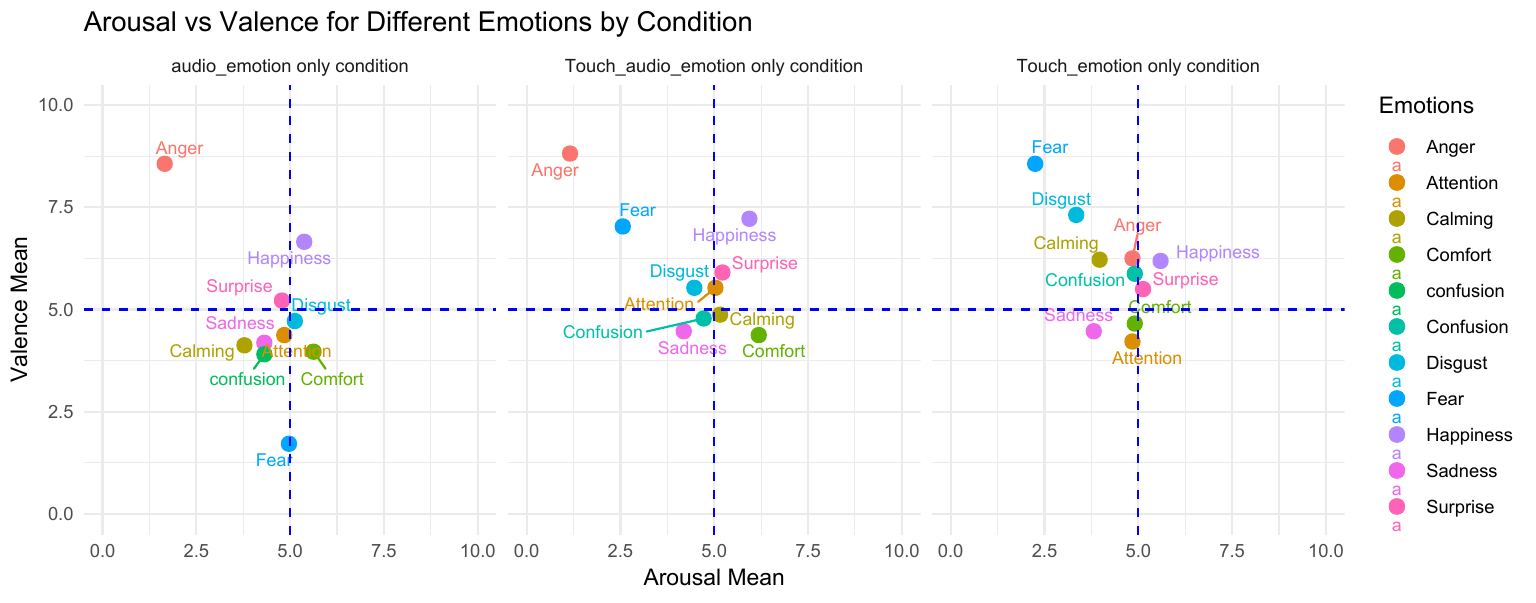} \caption{Decoding performance across different modalities. The figure compares unimodal (tactile or auditory) and multimodal (tactile + auditory) approaches, showing that multimodal integration consistently outperforms single-modality decoding.} \label{Decoding_result} 
\end{figure*}


\begin{figure}[h]
    \centering
        \includegraphics [width=\columnwidth]{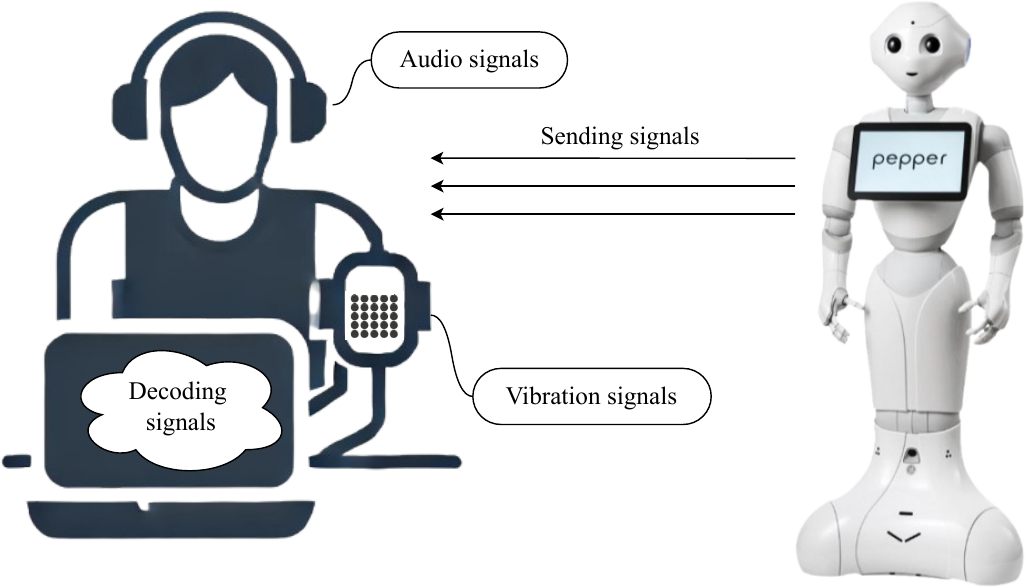}
    \caption{Mediate touch from the robot.}
    \label{interaction}
\end{figure}

\label{sec2:Materials}

\subsection{Equipment}

\begin{figure*}
\centering
\begin{subfigure}{0.49\textwidth}
    \centering
    \includegraphics[height=4cm]{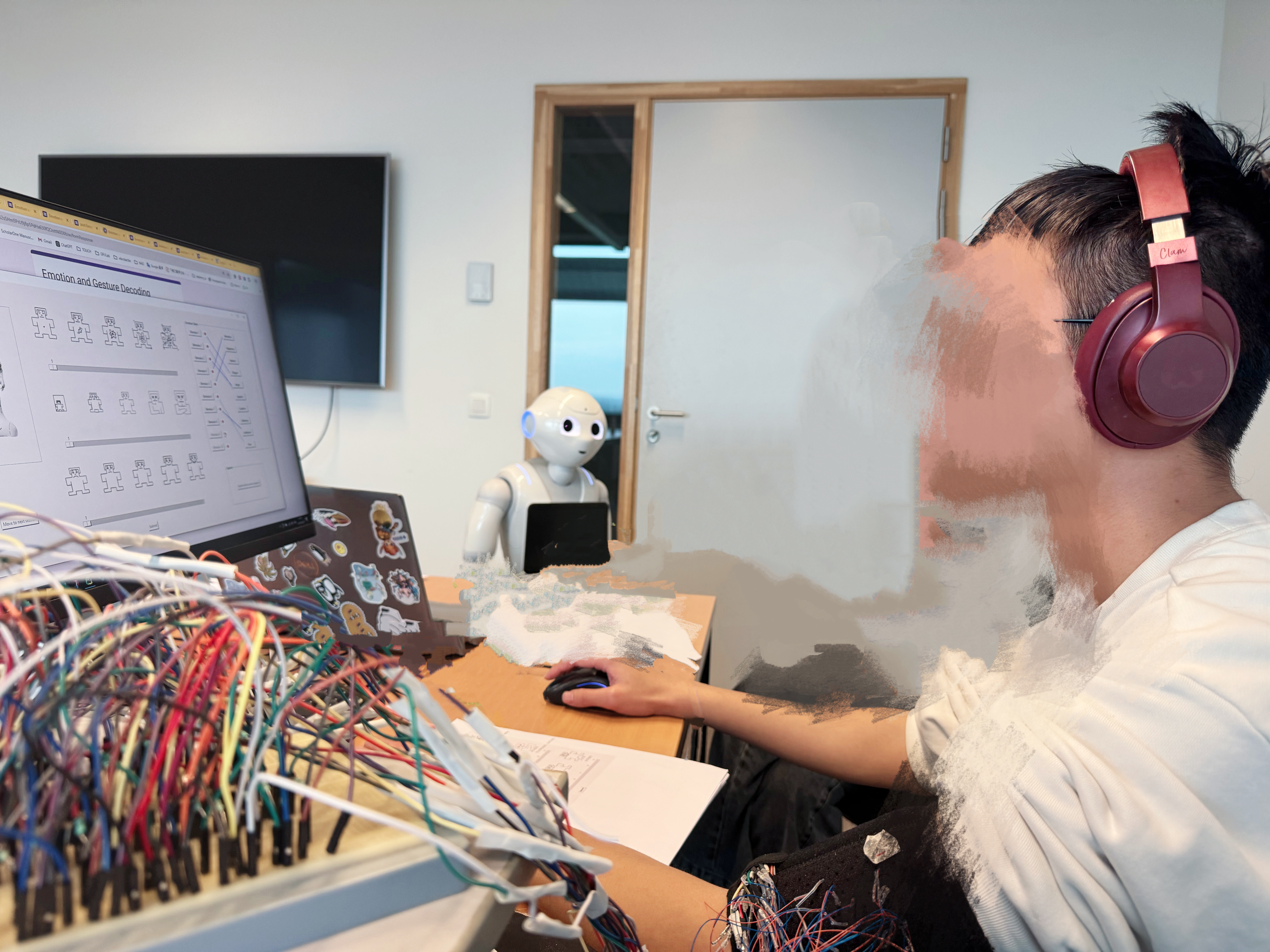}
\caption{Participants decoding emotions.}
    \label{fig:interaction}
\end{subfigure}
\begin{subfigure}{0.49\textwidth}
    \centering
    \includegraphics[height=4cm]{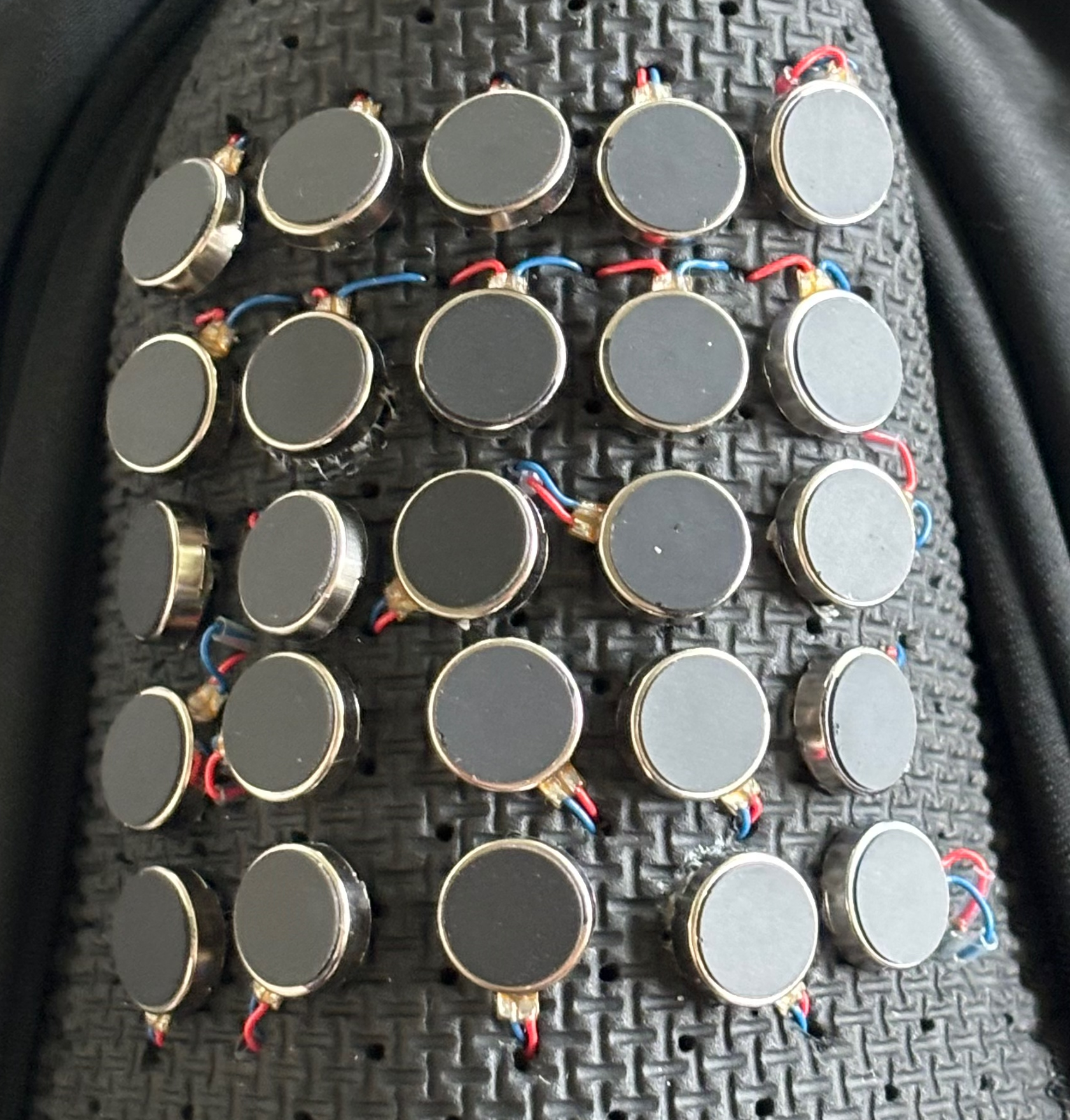}
    \caption{Distribution of vibration motors.}
    \label{fig:motors}
\end{subfigure}
\caption{Vibration sleeves.}
\label{fig:experimentalsetup}
\end{figure*}


\subsubsection{Vibration sleeve}

Many previous vibration-based haptic devices treat each motor as an independent stimulation point, focusing primarily on conveying discrete spatial locations or symbolic information. For example, large-spacing air actuator arrays and garments such as the ``Hug Shirt'' \cite{cheok2019huggy} have been designed to activate individual actuators to represent specific points or patterns on the body. Other designs use a single vibration motor to examine how vibration characteristics—such as frequency, amplitude, and duration—affect perceived arousal and valence \cite{ju2021haptic}. While these approaches are effective for delivering spatial cues or studying isolated perceptual parameters, they often lack the continuity and nuanced spatiotemporal dynamics that are characteristic of natural human touch.

In contrast, our design aims to move towards the relative and continuous qualities of human touch, producing tactile patterns that feel more organic and socially expressive rather than isolated point stimulations. To achieve this, we implemented a vibration device that consists of a 5x5 grid of vibration motors (6.5 centimetres(cm) $\times$ 6.5cm in total) embedded in an upper arm sleeve, powered by a 3.3V power supply. Each motor is controlled by a Raspberry Pi using BC557B PNP transistors, which act as switches to regulate current flow to the motors, as shown in Fig.~\ref{fig:motors} and Fig.~\ref{fig:interaction}. The system uses pulse-width modulation (PWM) for precise control over the intensity of each motor’s vibration, which allows for modulation of the vibration intensity by rapidly switching the transistors on and off at varying duty cycles. By adjusting the duty cycle of the PWM signal, the perceived intensity of each motor’s vibration can be finely controlled, allowing for a dynamic range of vibration patterns. 

\subsubsection{Headphone}

To prevent participants from hearing the sound of the vibrations during the experiment, we required them to wear active noise-cancelling headphones throughout the session to block out noise or any auditory cues generated by the motor vibrations.

\subsubsection{Questionnaire}

We collected subjective feedback through a questionnaire with four questions aimed at understanding the challenge of interpreting affective touch expression and the different modalities' contribution to emotion decoding. As shown below:

\begin{enumerate}
    \item \textit{Difficulty in Decoding Emotions:} We asked participants which emotions or intentions they find most challenging to interpret through a single multiple-choice question: ``Which emotion was the most difficult to decode?
    
    \item \textit{Confidence in Decoding Emotions Across Modalities:} Participants rated their confidence in decoding emotions through different sensory channels: ``How confident were you in decoding emotions using vibrations?”; ``How confident were you in decoding emotions using sound?”; and ``How confident were you in decoding emotions using both sound and vibration?” (Likert scale: 1 = Not confident at all, 5 = Very confident).
    
    \item \textit{Primary Sensory Channel Preference:} Participants were asked to identify the sensory channel they primarily relied on to decode emotions. Options included the sound modality, the touch modality, or a combination of both (multiple choice).

    \item \textit{Emotion Decoding Strategy:} Participants were asked to indicate the specific strategies they used to decode emotions, with the option to select multiple strategies. These included: relying on vibration characteristics (e.g., intensity, magnitude, frequency, speed), interpreting sound cues such as volume or pitch, assessing the gesture’s speed or rhythm, using a combination of these features, recognizing the type of touch gesture experienced, selecting all of the above, or relying purely on intuitive or gut feeling.

    \item \textit{Confusing Emotions or Gestures:} An optional question asked participants if there were any particular emotions or gestures that were especially confusing.

\end{enumerate}

\subsection{Dataset}

\begin{figure}[h]
    \centering
        \includegraphics [height=10cm]{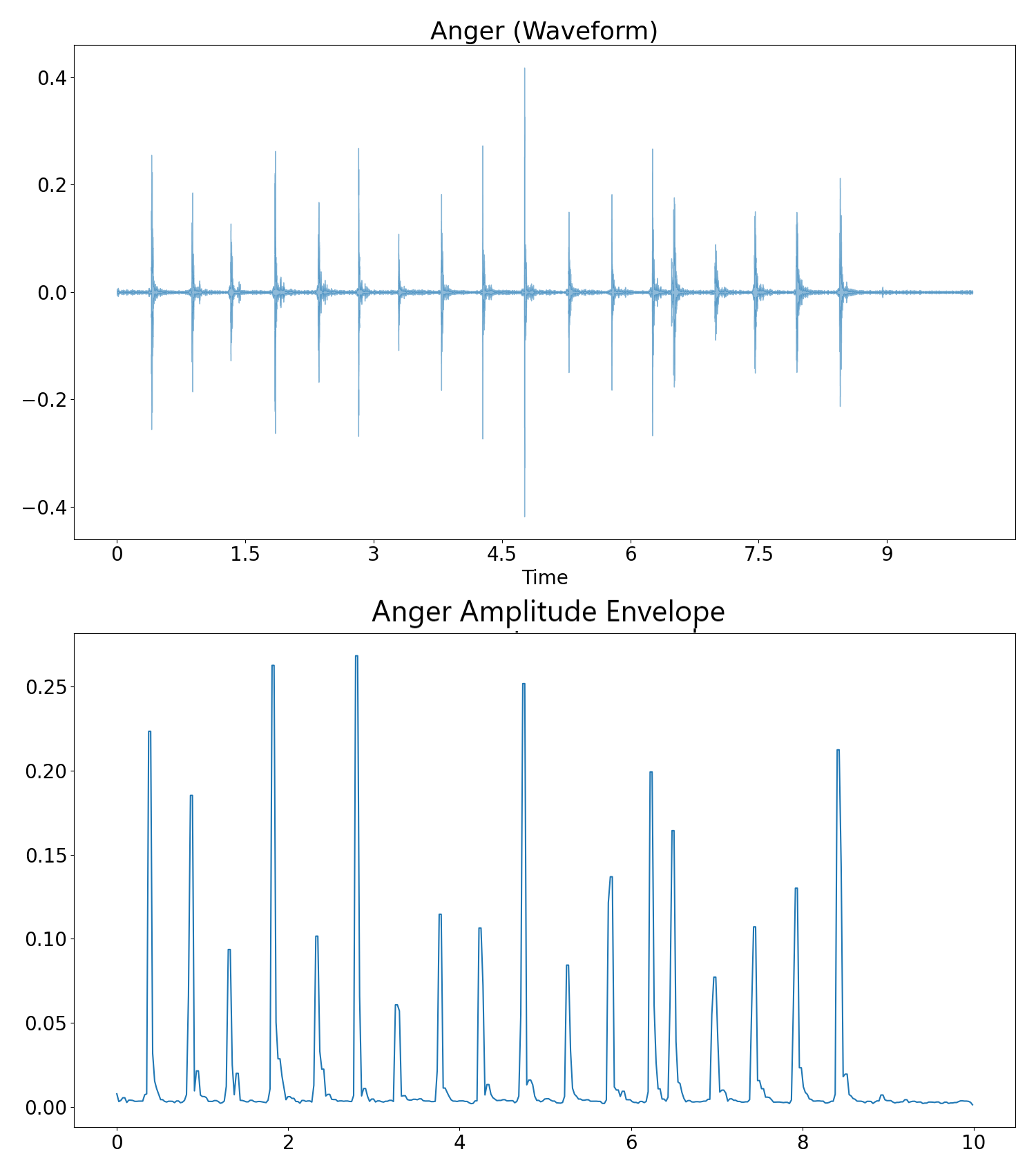}
    \caption[Recording of ``Anger'' stimulus.]{Top: Waveform of ``Anger'' audio, showing the raw time-domain signal. Bottom: Original amplitude envelope of the same excerpt, illustrating the smoothed contour of overall loudness variations across time.}
    \label{fig::anger}
\end{figure}

\begin{figure}[h]
    \centering
        \includegraphics [height=10cm]{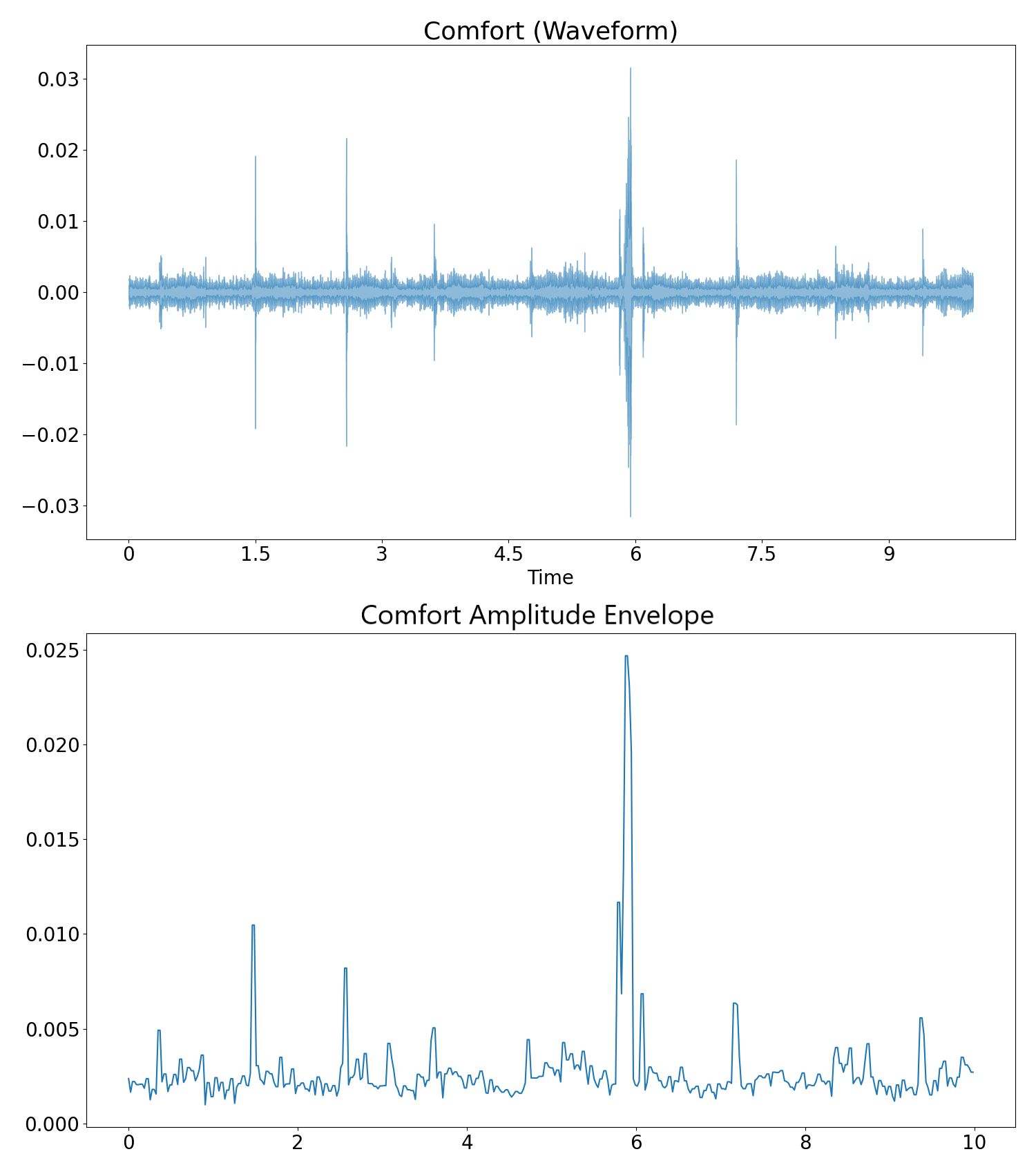}
    \caption[Recording of ``Comfort'' stimulus.]{Top: Waveform of the ``Comfort'' audio excerpt in the time domain. Bottom: Original amplitude envelope of the same excerpt, highlighting the smoothed variations in loudness across time.}
    \label{fig::comfort}
\end{figure}

We conducted data collection experiments to separately record touch sounds associated with gestures and those associated with emotions, as described in \cite{ren2024conveying}. Participants first expressed six distinct gestures on the left arm of the Pepper robot and subsequently expressed ten different emotions through similar touch interactions.The sounds generated by these interactions were recorded, resulting in a dataset of 84 audio clips (28 participants $\times$ 3 rounds), sampled at 44.1\,kHz. Examples of the recordings for ``Anger'' and ``Comfort'' are shown in Fig.~\ref{fig::anger} and Fig.~\ref{fig::comfort}, respectively. In addition, 84 tactile files were collected in CSV format at 45\,Hz, each with a duration of 10\,s, covering both gestures and emotions. Examples of combined auditory and haptic cue visualizations, along with the recordings, are available at the GitHub repository:  \footnote{\url{https://github.com/qiaoqiao2323/Multi_vibro/tree/main}}.

To analyse this data, we applied the k-means clustering algorithm, grouping the audio features and tactile features in \cite{ren2024conveying} into three clusters based on their acoustic features. We identified the most populated cluster, representing the largest grouping of participants, as the most characteristic. Within this cluster, we calculated the Euclidean distance of each participant's expression features to the cluster centroid (average feature values) to assess each clip's representativeness. The three participants' IDs with the smallest distances to the centroid were selected as the most representative, and from these, we chose one participant's data among the three to serve as a representative stimulus for all participants for each emotion. Then we translate the tactile files to vibrations, each vibration stimulus lasts for 10 seconds as well. By adjusting the duty cycle of the PWM signal, the perceived intensity of each motor’s vibration can be finely controlled, allowing the replication of tactile dynamic patterns and converting them to vibration patterns.

\section{Methods}

\subsection{Participants}

Thirty-two Chinese participants (17 female, 15 male; M = 27.8, SD = 2.3 years) participated in the experiment. Participants were recruited from a similar cultural background to ensure consistency in emotional interpretation. The study complied with ethical guidelines established by \emph{Ghent University}, and informed consent was obtained from all. The entire experiment lasts for one hour, and each participant receives 10 euros as compensation for their participation. In addition, to maintain engagement, the participant with the highest score receives a 30 euro reward. The ranking is presented anonymously to all participants.

\subsection{Procedures}

\begin{figure}[h]
    \centering
        \includegraphics [height=3.9cm]{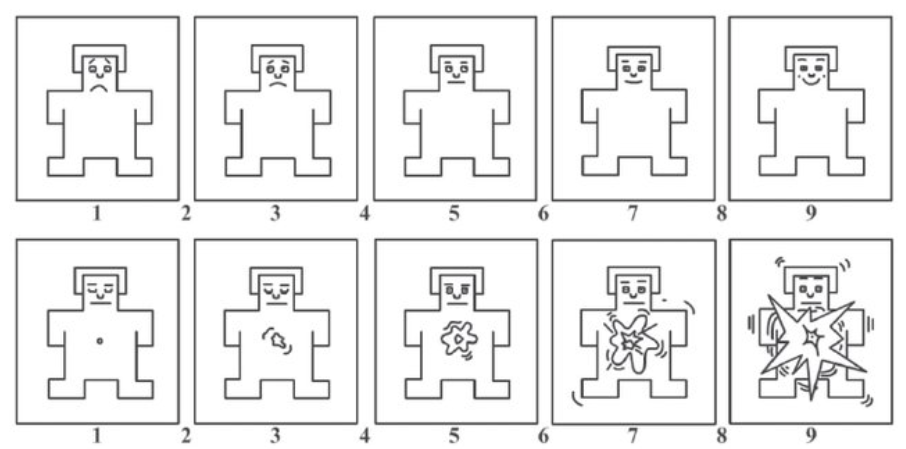}
    \caption[SAM for valence and arousal.]{SAM for valence and arousal, the first row is valence and the second is arousal.}
    \label{fig::sam_}
\end{figure}

Each participant underwent four phases as shown in Fig.~\ref{interaction} and Fig.~\ref{Decoding_result}; in each phase, they were asked to decode emotions for each stimulus by Self Assessment Manikins (SAM) as shown in Fig.~\ref{fig::sam_}.

\vspace{-3mm}

\subsection{Experimental design}

To explore emotion decoding via touch-based expression and reveal how consistently emotions and gestures are expressed across different individuals, we set up a data collection experiment, which can be seen in Fig.~\ref{fig:experimentalsetup}. The experiment followed a within-subject design using three sessions, including a pre-session, a touch gesture decoding session, and an emotion decoding session. During each session, participants are allowed to replay the stimulus multiple times.

\begin{enumerate}

    \item \textit{Pre-Session:} In this preliminary session, participants are asked if they can feel the vibration to ensure it meets their threshold for tactile perception. If necessary, the amplitude of the stimulus is adjusted.

    \item \textit{Study 1: Emotion Decoding Session:}
    \begin{itemize}
        \item \textit{Touch Modality:} Participants decode emotions conveyed through vibration, rating arousal and valence for each stimulus, and forced-choice selecting the emotion from a list of ten options.
        \item \textit{Sound Modality:} Participants decode emotions conveyed through sound, rate the arousal and valence for each stimulus, and identify the emotion from the provided list. During this session, the sound level for each stimulus is played at the same level.
        \item \textit{Combined Touch and Sound Modality:} Participants decode emotions using synchronized sound and vibration, rate arousal and valence, and identify the emotion from the list of ten options.
    \end{itemize}

    \item \textit{Study 2: Touch Gestures Decoding Session:}
    \begin{itemize}
        \item \textit{Touch Modality:} Participants decode touch gestures using vibration alone, rating the arousal and valence for each stimulus, and identifying the gesture from a list of provided options.
        \item \textit{Sound Modality:} Participants decode touch gestures using sound, rate arousal and valence for each stimulus, and identify the gesture from the provided options. During this session, the sound level for each stimulus is at the same level.
        \item \textit{Combined Touch and Sound Modality:} Participants decode touch gestures using synchronized sound and vibration, rate the arousal and valence, and select the corresponding gesture from the provided list.
    \end{itemize}

\end{enumerate}







\section{Results and analysis}

\subsection{Sound modality}

\subsubsection{Emotion decoding}

We calculated the average arousal and valence ratings provided by participants for various emotional stimuli. The results for the 10 emotions are summarized in Table.~\ref{tab:emotion_va_sound}, and their distribution within Russell's circumplex model is depicted in Figure~\ref{russel_model}. Participants placed ``Happiness'' in the high-arousal, positive-valence quadrant. Emotions such as ``Surprise'' and ``Anger'' were placed in the high-arousal, negative-valence quadrant. ``Confusion'', ``Sadness'', ``Calming'', ``Fear'', and ``Attention'' were associated with the low-arousal, negative-valence quadrant; while ``Comfort'' and ``Disgust'' were classified in the low-arousal, positive-valence quadrant.

This classification generally aligns with the emotion categorization described in Russell's circumplex model and findings from previous studies \cite{noordewier2013valence, bradley2001emotion}. However, discrepancies were observed. For instance, ``Calming'' is typically expected to fall in the low-arousal, positive-valence quadrant but was misclassified into the low-arousal, negative-valence quadrant. Participants mentioned they were confused ``Comfort'', ``Calming'', and ``Sadness'' when they tried to decode emotions. Similarly, ``Fear'' was misclassified into the low-arousal, negative-valence quadrant. A possible explanation is that participants expressed fear by tightly grasping the robot's arm and shaking it rapidly, which produced minimal sound but a large force, which might lead to the wrong classification only based on the sound channel.

Additionally, ``Disgust'' was misclassified into the low-arousal, positive-valence quadrant. Participants expressed difficulty decoding ``Disgust''. Observational data indicate that participants often used a single finger to push the robot away multiple times. Since this action was performed slowly, the resulting sound was perceived as gentle and deliberate, despite the actual force being significant. These findings highlight potential challenges in accurately categorizing emotions based on arousal and valence, particularly when relying on multi-sensory input.

As our findings showed, participants often misclassified emotions that fell within the same emotional quadrant. To better understand this pattern, we analyzed decoding accuracy across four quadrants plus origin based on arousal and valence dimensions. The results indicated that participants were generally able to distinguish emotions across different quadrants, suggesting that the sound modality effectively conveys differences in arousal and valence. However, decoding became more difficult when emotions were closer in affective space—particularly those sharing the same arousal and valence levels. Interestingly, decoding accuracy for emotions in Quadrant 2 (high arousal, negative valence) was higher than in other quadrants. One possible explanation is that high-arousal negative stimuli tend to have distinctive acoustic features, such as large sound amplitude or high frequency, which participants may associate more consistently with this emotional zone.

\begin{table*}\footnotesize
\setlength{\abovecaptionskip}{0.0cm}   
	\setlength{\belowcaptionskip}{-0cm}  
	\renewcommand\tabcolsep{2.0pt} 
	\centering
	\caption{Arousal and valence of different emotions and decoding accuracy(\%) based on sound modality, decoding accuracy for each emotion was compared against the chance level (10\%, corresponding to random guessing among ten classes). A one-sample binomial test was used to determine whether the observed decoding accuracy was significantly higher than chance. The reported p-values indicate the results of this comparison, with significant values shown in bold.}
	\begin{tabular}
	{
	p{1.5cm}<{\centering} 
 p{1cm}<{\centering} 
	 p{1.5cm}<{\centering} 
  p{1.5cm}<{\centering}
	p{1cm}<{\centering}
 	p{1.5cm}<{\centering}
        p{1cm}<{\centering}
 	p{1.5cm}<{\centering}
 	p{1.5cm}<{\centering}
 	p{1.5cm}<{\centering}
 	p{1cm}<{\centering}
	} 
\hline

     {Emotions} & {Happiness} & {Surprise}  & {Fear} & {Disgust} & {Anger} & {Comfort}  & {Attention} & {Calming} & {Confusion} & {Sadness} \\

\hline
   {Arousal} & {$6.7\pm1.4$} & {$5.2\pm2.1$} & {$1.7\pm1.3$} & {$4.7\pm1.8$} & {$8.6\pm1.5$} & {$4.0\pm1.8$} & {$4.4\pm1.9$} & {$4.1\pm2.1$} & {$3.9\pm2.1$} & {$4.2\pm1.6$}\\
   {Valence} & {$5.4\pm1.8$} & {$4.8\pm2.2$} & {$5.0\pm1.6$} & {$5.1\pm2.0$} & {$1.7\pm1.6$} & {$5.6\pm2.3$} & {$4.8\pm1.9$} & {$3.8\pm1.9$} & {$4.3\pm2.0$} & {$4.3\pm1.8$}\\
    {Accuracy} & {$\textbf{34.4}$} & {$21.9$} & {$21.9$} & {$\textbf{34.4}$} & {$\textbf{81.2}$} & {$\textbf{25.0}$} & {$\textbf{43.8}$} & {$12.5$} & {$21.9$} & {$18.8$}\\
    {Sig.(p)} & {$<0.01$} & {$0.06$} & {$0.06$} & {$<0.01$} & {$<0.01$} & {$0.03$} & {$<0.01$} & {$0.34$} & {$0.06$} & {$0.11$}\\

     \hline

    \end{tabular}
\label{tab:emotion_va_sound}
\end{table*}

\begin{figure}
    \centering
    \includegraphics[width=\columnwidth]{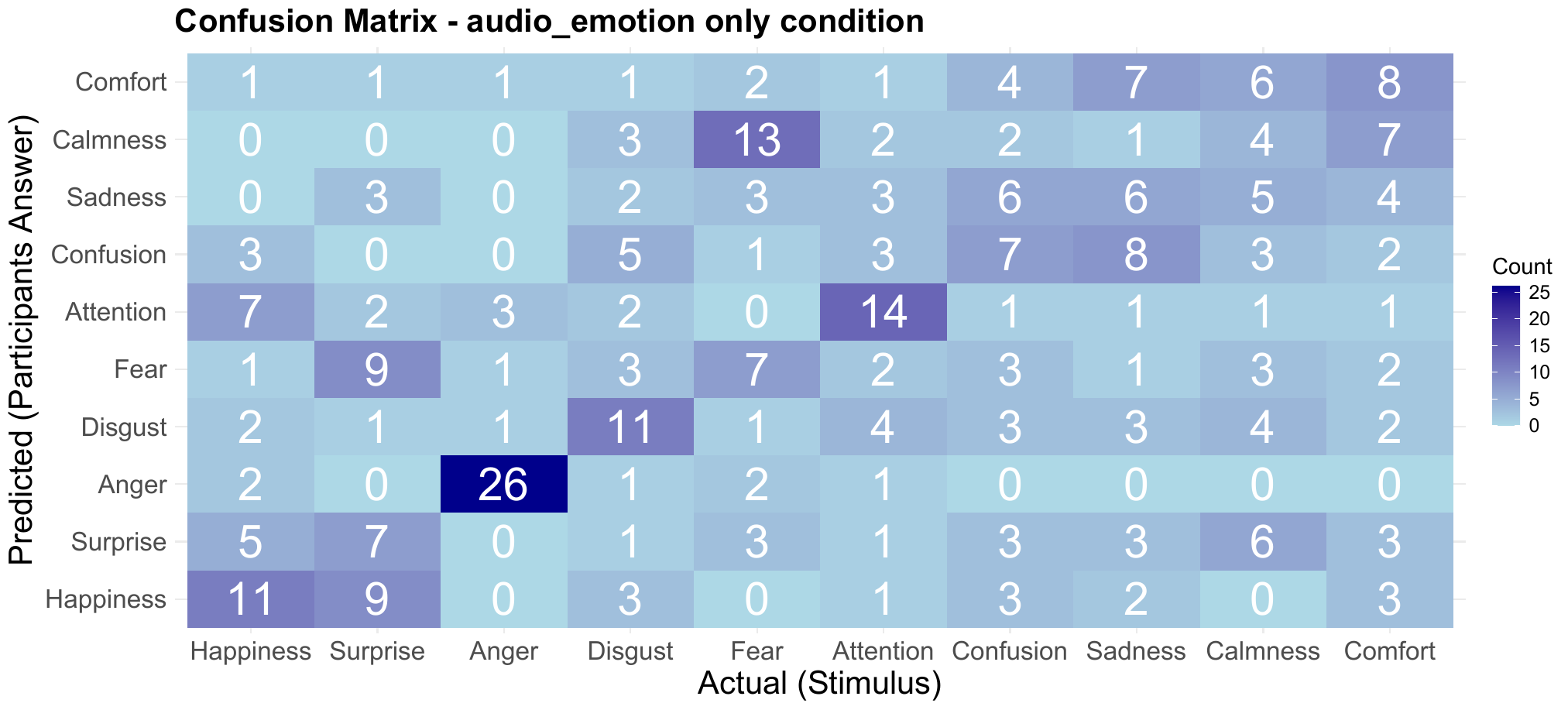}
    \caption{Confusion matrix for emotions (sound modality).}
    \label{fig:emotion_confusion_sound}
\end{figure}



The pair decoding results in Fig.~\ref{fig:emotion_confusion_touch} showed that 26 out of 32 participants correctly identified ``Anger'', making it the most accurately decoded emotion. This was followed by ``Attention'', which 14 participants decoded correctly. We tested whether participants' emotion decoding performance was significantly above chance (10\%) for sound modality using binomial tests. The decoding accuracy was 31.6\% across 10 emotions, above chance, $p < 0.01$, 95\% CI [0.273, 1].

\subsubsection{Touch gesture decoding}

As shown in the Table.~\ref{tab:gesture_audio} and Fig.~\ref{fig:gesture_confusion_touch}, the participants' touch gesture decoding accuracy is $66.1\%$; specifically, the gesture that participants found easiest to decode was ``Tickle'', followed by ``Rub''. In contrast, ``Poke'', ``Pat'' and ``Tap'' were more frequently misclassified and often confused with each other. A likely reason is that these gestures produce acoustically similar, short percussive events with overlapping temporal rhythms and spectral envelopes, which are difficult to differentiate using audio alone. By comparison, ``Tickle'' (irregular, high‑frequency friction) and ``Rub'' (sustained friction) provide more distinctive acoustic cues. Overall, all gestures were decodable from sound, and gesture decoding outperformed emotion decoding.

\begin{figure}
    \centering
    \includegraphics[height=4.6cm]{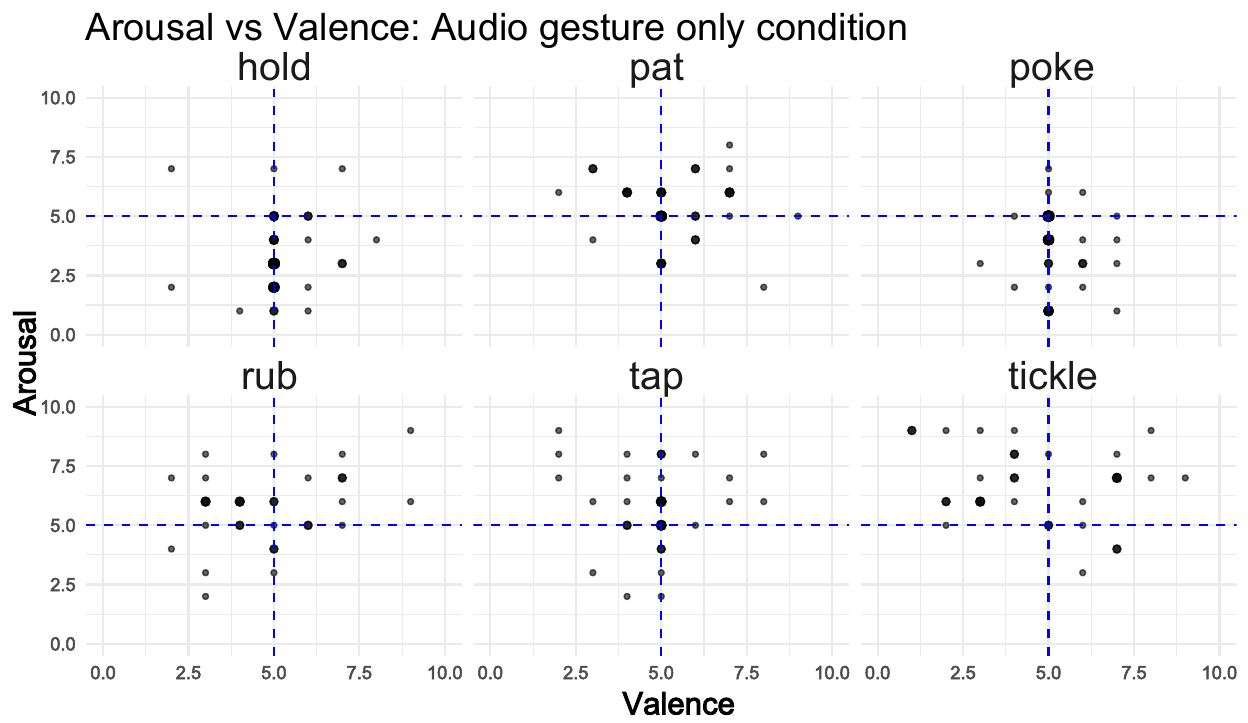}
    \caption{Scatter plots for affective response of touch gestures in the sound modality.}
    \label{fig:scatter_gesture_audio}
\end{figure}

Regarding the affective responses to touch gestures based on the sound modality (as shown in the Figure.~\ref{fig:scatter_gesture_audio} and Table.~\ref{tab:gesture_audio}), the average valence ratings across all gestures were generally close to neutral (around 5). However, arousal levels varied more distinctly between different gestures, suggesting that arousal may play a more discriminative role. Participants tended to perceive “Pat” as slightly positive and moderately arousing, while “Hold” and “Poke” were associated with low arousal and slightly positive valence. In contrast, ``Rub”, ``Tap” and ``Tickle” were rated as higher in arousal and also slightly more positive in valence.

\begin{figure}
    \centering
    \includegraphics[width=\columnwidth]{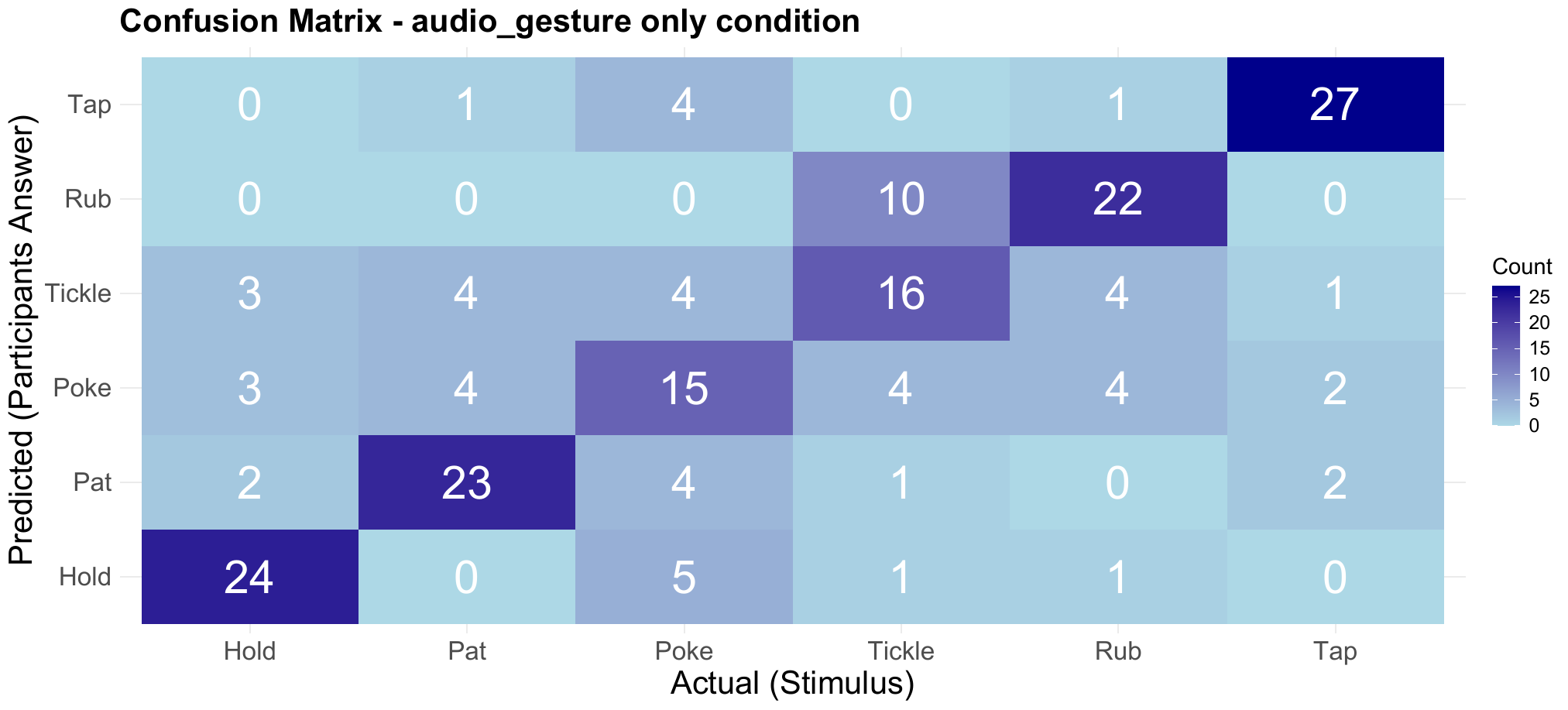}
    \caption{Confusion matrix for touch gestures (sound modality).}
    \label{fig:gesture_confusion_sound}
\end{figure}


\begin{table}
\footnotesize
\setlength{\abovecaptionskip}{0.0cm}   
	\setlength{\belowcaptionskip}{0.0cm}  
	\renewcommand\tabcolsep{2.0pt} 
	\centering
	\caption{Decoding accuracy(\%) based on sound modality. Decoding accuracy for each quadrant was compared against the chance level (20\%, corresponding to random guessing among four quadrants plus the origin). A one-sample binomial test was used to determine whether the observed accuracy was significantly above chance. The reported p-values indicate significance level, with the highest decoding results in bold.}
	\begin{tabular}
	{
	p{2.5cm}<{\centering} 
 p{1cm}<{\centering} 
	 p{1.5cm}<{\centering} 
  p{1cm}<{\centering}
	p{1cm}<{\centering}
 	p{1cm}<{\centering}
	} 
\hline
      
     {Quadrant Accuracy} & {Q0} & {Q1}  & {Q2} & {Q3} & {Q4}  \\
     
\hline
   {SM Accuracy} & {$43.8$} & {$50.0$} & {$\textbf{55.2}$} & {$42.2$} & {$39.1$}  \\
   {Sig.(p)} & {$<0.01$} & {$<0.01$} & {$<0.01$} & {$<0.01$} & {$<0.01$} \\
   \hline
      {TM Accuracy} & {$34.4$} & {$45.3$} & {$\textbf{61.5}$} & {$35.9$} & {$34.4$}  \\
    {Sig.(p)} & {$<0.01$} & {$<0.01$} & {$<0.01$} & {$<0.01$} & {$<0.01$} \\
   \hline  
   {STM Accuracy} & {$53.1$} & {$53.1$} & {$\textbf{74.0}$} & {$42.2$} & {$56.2$}  \\
  {Sig.(p)} & {$<0.01$} & {$<0.01$} & {$<0.01$} & {$<0.01$} & {$<0.01$} \\
   \hline

    \end{tabular}
\label{tab:gesture_quadrant}
\end{table}

      



\begin{table}
\footnotesize
\setlength{\abovecaptionskip}{0.0cm}   
	\setlength{\belowcaptionskip}{-0cm}  
	\renewcommand\tabcolsep{2.0pt} 
	\centering
	\caption{Decoding accuracy(\%) and affective response of touch gestures based on sound modality. Decoding accuracy for each gesture was compared against the chance level (16.7\%, corresponding to random guessing among six classes). A one-sample binomial test was used to determine whether the observed accuracy was significantly higher than chance. The reported p-values indicate the results of this comparison.}
	\begin{tabular}
	{
	p{1.7cm}<{\centering} 
 p{0.9cm}<{\centering} 
	 p{1.4cm}<{\centering} 
  p{0.9cm}<{\centering}
	p{0.9cm}<{\centering}
 	p{0.9cm}<{\centering}
        p{0.9cm}<{\centering}
	} 
\hline
      
     {Gestures} & {Hold} & {Pat}  & {Poke} & {Rub} & {Tap} & {Tickle}  \\
     
\hline
   {Accuracy} & {$75.0$} & {$71.9$} & {$46.9$} & {$68.8$} & {$84.4$} & {$50.0$} \\
   {Sig.(p)} & {$<0.01$} & {$<0.01$} & {$<0.01$} & {$<0.01$} & {$<0.01$} & {$<0.01$} \\
   \hline
   {Arousal\_mean} & {$3.38$} & {$5.34$} & {$3.62$} & {$5.72$} & {$5.81$} & {$6.72$} \\
    {Arousal\_std} & {$1.72$} & {$1.38$} & {$1.64$} & {$1.57$} & {$1.80$} & {$1.63$} \\
    \hline
    {Valence\_mean} & {$5.22$} & {$5.41$} & {$5.28$} & {$4.81$} & {$4.78$} & {$4.66$} \\
    {Valence\_std} & {$1.18$} & {$1.54$} & {$0.89$} & {$1.87$} & {$1.48$} & {$2.21$} \\

     \hline

    \end{tabular}
\label{tab:gesture_audio}
\end{table}

\subsection{Touch modality}

\begin{figure}
    \centering
    \includegraphics[width=\columnwidth]{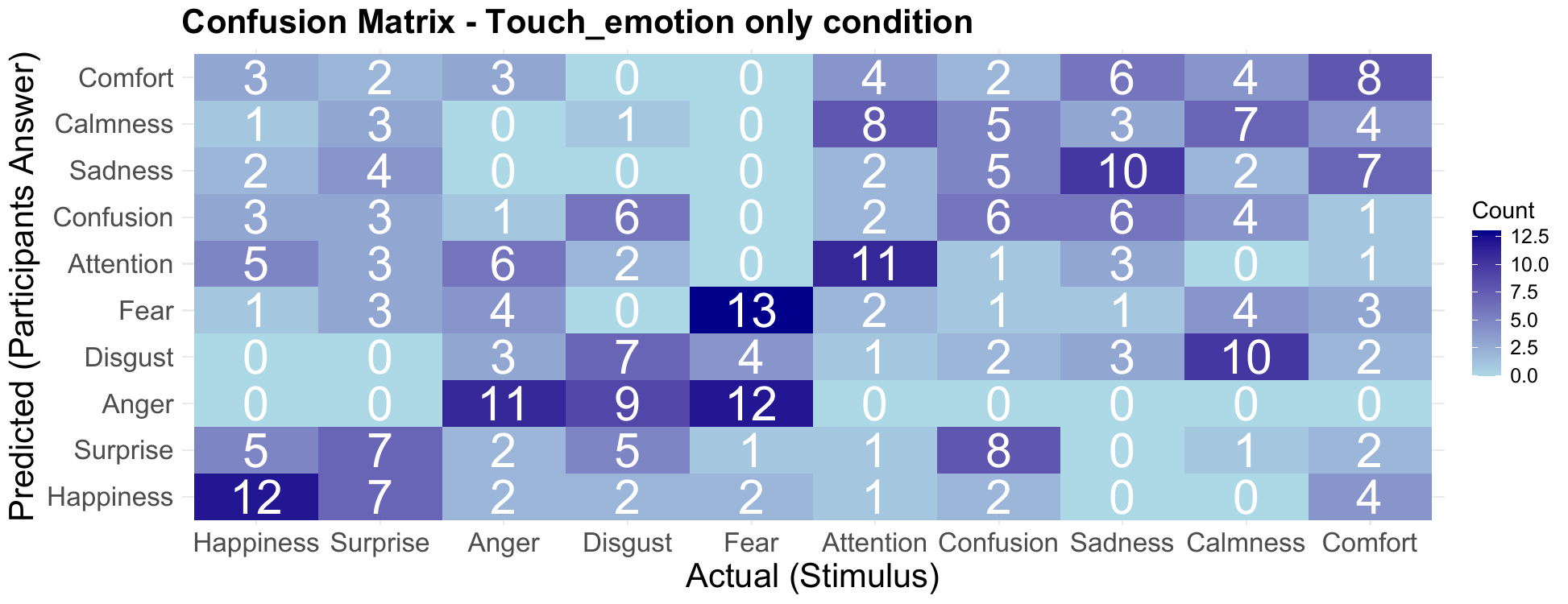}
    \caption{Confusion matrix for emotions (touch modality).}
    \label{fig:emotion_confusion_touch}
\end{figure}

We averaged the participants' arousal and valence ratings for different emotional stimuli. The results for the 10 emotions are shown in Table.~\ref{tab:emotion_both}. Figure.~\ref{fig:emotion_confusion_touch} illustrates the distribution of the 10 emotions within Russell's circumplex model.

\begin{table*}\footnotesize
\setlength{\abovecaptionskip}{0.0cm}   
	\setlength{\belowcaptionskip}{-0cm}  
	\renewcommand\tabcolsep{2.0pt} 
	\centering
	\caption[Arousal and valence of different emotions and decoding accuracy(\%) based on touch modality.]{Arousal and valence of different emotions and decoding accuracy(\%) based on touch modality. For each emotion, decoding accuracy was evaluated relative to the chance level (10\%). A binomial proportion test was applied to examine whether performance was significantly above random guessing. The resulting p-values are reported, with statistically significant effects indicated in bold.}
	\begin{tabular}
	{
	p{1.5cm}<{\centering} 
 p{1cm}<{\centering} 
	 p{1.5cm}<{\centering} 
  p{1.5cm}<{\centering}
	p{1cm}<{\centering}
 	p{1.5cm}<{\centering}
        p{1cm}<{\centering}
 	p{1.5cm}<{\centering}
 	p{1.5cm}<{\centering}
 	p{1.5cm}<{\centering}
 	p{1cm}<{\centering}
	} 
\hline

     {Emotions} & {Happiness} & {Surprise}  & {Fear} & {Disgust} & {Anger} & {Comfort}  & {Attention} & {Calming} & {Confusion} & {Sadness} \\

\hline
   {Arousal} & {$6.2\pm2.1$} & {$5.5\pm2.3$} & {$8.6\pm1.3$} & {$7.3\pm1.9$} & {$6.3\pm1.8$} & {$4.7\pm2.3$} & {$4.2\pm2.1$} & {$6.2\pm2.0$} & {$5.9\pm2.0$} & {$4.5\pm2.1$}\\
   {Valence} & {$5.6\pm2.2$} & {$5.1\pm2.5$} & {$2.3\pm2.5$} & {$3.3\pm2.5$} & {$4.8\pm2.4$} & {$4.9\pm1.8$} & {$4.8\pm1.8$} & {$4.0\pm2.1$} & {$4.9\pm2.1$} & {$3.8\pm1.7$}\\
    {Accuracy} & {$\textbf{34.4}$} & {$\textbf{34.4}$} & {$\textbf{40.6}$} & {$18.8$} & {$18.8$} & {$\textbf{25.0}$} & {$\textbf{34.4}$} & {$21.9$} & {$18.8$} & {$15.6$}\\
    {Sig.(p)} & {$<0.01$} & {$<0.01$} & {$<0.01$} & {$0.11$} & {$0.11$} & {$0.03$} & {$<0.01$} & {$0.06$} & {$0.11$} & {$0.20$}\\

     \hline
    \end{tabular}
\label{tab:emotion_both}
\end{table*}


\begin{figure}
    \centering
    \includegraphics[width=\columnwidth]{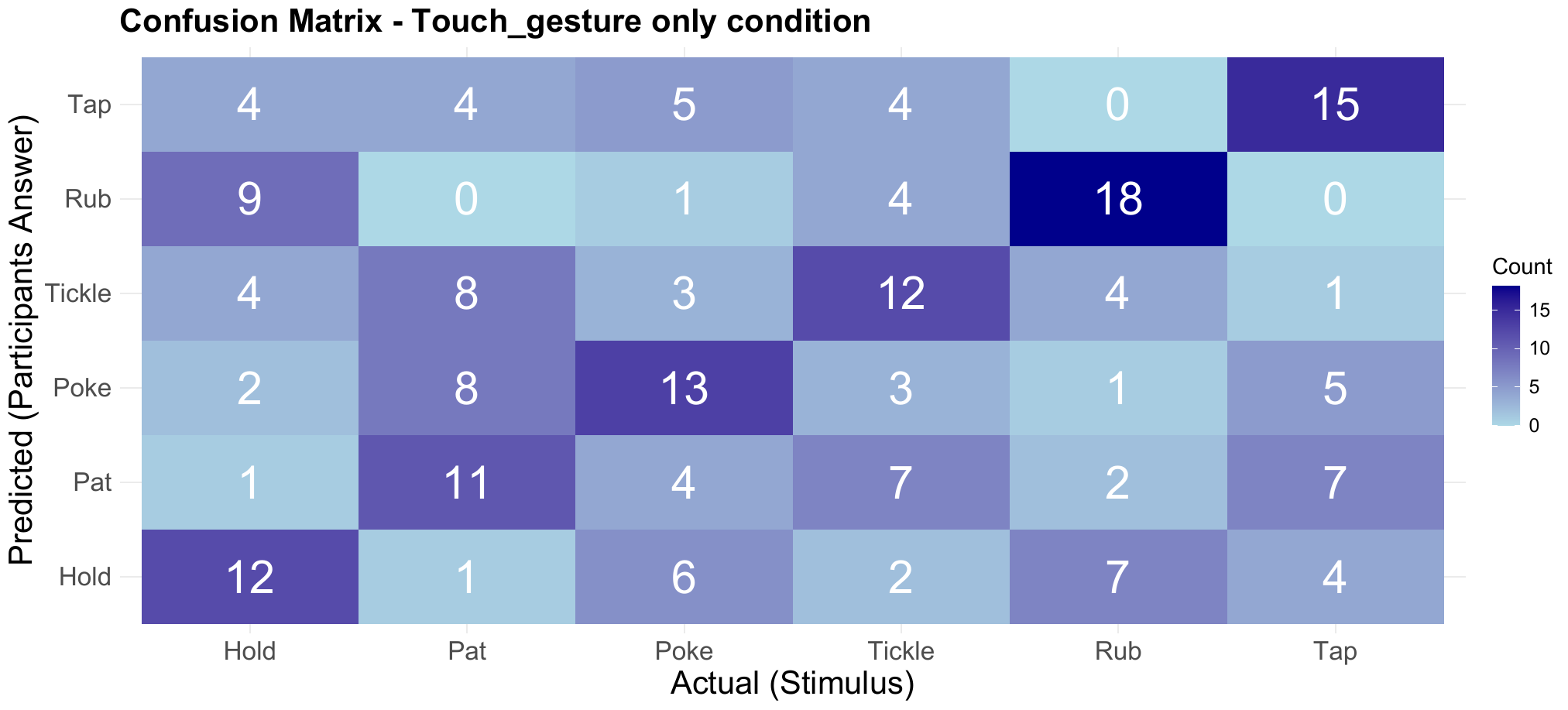}
    \caption{Confusion matrix for touch gestures (touch modality).}
    \label{fig:gesture_confusion_touch}
\end{figure}

Participants placed ``Happiness'' and ``Surprise'' in the high-arousal, positive-valence quadrant. ``Disgust, ``Fear'',  ``Calming'', ``Confusion'' and ``Anger'' were categorized in the high-arousal, negative-valence quadrant. ``Sadness'', ``Attention'' and ``Comfort'' were assigned to the low-arousal, negative-valence quadrant, while no emotion was classified in the low-arousal, positive-valence quadrant. Some decoding classifications are consistent with the emotion categorization described in Russell's circumplex model and previous research, while some emotions like ``Calming'' and ``Confusion'' are misclassified to high arousal and negative quadrant, plus ``'Attention'' is misclassified to low arousal, and negative valence quadrant, which might be some information missed when transferring the information, for example, participants tend to hold the robot's arm to convey ``Calming'', however, it leads to strong vibrations, which might mistakenly convey conflicting cues. Additionally, ``Anger'' was frequently placed near the neutral zone, which might be due to hardware limitations, such as the size of the vibrosleeves and the motor updating frequency. Participants used fast, high-frequency pouching gestures to express anger, but the robot’s vibration motors require a certain time to start and stop. This latency likely resulted in short and weak vibration bursts, reducing the perceived intensity of the emotion and potentially leading to its misclassification.

The overall decoding accuracy across the 10 emotions is 25\%, and the pair decoding results in Fig.~\ref{fig:emotion_confusion_sound} showed that 13 out of 32 participants correctly identified ``Fear'', making it the most accurately decoded emotion. This was followed by ``Happiness'' and ``Attention'', which both were correctly decoded by 11 participants. The decoding performance was notably lower than that of the sound modality, which may be attributed to individual differences in tactile sensitivity. While the human auditory system is finely tuned to detect a wide range of sound frequencies, the skin's sensitivity to vibrations is more limited and varies across individuals and body regions \cite{sakaguchi2023dynamic}. These sensory differences likely contribute to the lower accuracy observed in the touch-based condition. Despite this, participants were still able to successfully recognize emotions across different arousal–valence quadrants using the touch modality, as shown in the Table.~\ref{tab:gesture_quadrant}, suggesting that affective touch can convey meaningful emotional information based on touch modality, particularly when the emotions differ in their overall arousal.



As shown in Table.~\ref{tab:gesture_touch} and Fig.~\ref{fig:gesture_confusion_touch}, the gesture most easily decoded by participants was ``Rub'', followed by ``Tap'' and ``Poke''. Additionally, ``Poke'', ``Pat'', and ``Tap'' were often misclassified and confused with one another. This is understandable, as ``Poke'' and ``Tap'' may share a similar contact area, while ``Pat'', ``Poke'', and ``Tap'' exhibit comparable movement rhythms. Similarly, ``Hold'' and ``Rub'' could be mistaken for each other due to their overlapping contact areas. The overall decoding accuracy is 42.2\%.

\begin{table}
\footnotesize
\setlength{\abovecaptionskip}{0.0cm}   
	\setlength{\belowcaptionskip}{-0cm}  
	\renewcommand\tabcolsep{2.0pt} 
	\centering
	\caption[Decoding accuracy(\%) and affective response of touch gestures based on touch modality.]{Decoding accuracy(\%) and affective response of touch gestures based on touch modality. For each gesture, decoding accuracy was evaluated relative to the chance baseline of 16.7\%. A binomial proportion test was applied to examine whether performance exceeded random guessing. The resulting p-values are reported.}
	\begin{tabular}
	{
	p{1.8cm}<{\centering} 
 p{1cm}<{\centering} 
	 p{1.5cm}<{\centering} 
  p{1cm}<{\centering}
	p{1cm}<{\centering}
 	p{1cm}<{\centering}
        p{1cm}<{\centering}
	} 
\hline
      
     {Gestures} & {Hold} & {Pat}  & {Poke} & {Rub} & {Tap} & {Tickle}  \\
     
\hline
   {Accuracy} & {$37.5$} & {$34.4$} & {$40.6$} & {$56.2$} & {$46.9$} & {$37.5$} \\
    {Sig.(p)} & {$0.01$} & {$0.02$} & {$<0.01$} & {$<0.01$} & {$<0.01$} & {$0.01$} \\
    \hline
    {Arousal\_mean} & {$7.28$} & {$6.66$} & {$7.66$} & {$6.47$} & {$5.97$} & {$4.97$} \\
    {Arousal\_std} & {$1.0$} & {$1.18$} & {$1.00$} & {$1.92$} & {$1.64$} & {$1.80$} \\
    \hline
    {Valence\_mean} & {$5.06$} & {$5.0$} & {$5.22$} & {$6.19$} & {$5.28$} & {$5.66$} \\
    {Valence\_std} & {$1.48$} & {$1.67$} & {$2.39$} & {$1.94$} & {$1.46$} & {$1.77$} \\

     \hline

    \end{tabular}
\label{tab:gesture_touch}
\end{table}

One possible explanation for this confusion is the limited size of the designed sleeve's touch-sensitive area. Although the 25-motor setup aims to facilitate relative haptic perception, the small touch area may hinder participants from distinguishing significant differences between certain gestures. For instance, while ``Poke'' involves a smaller contact area compared to ``Pat'', their rhythmic patterns can be similar, as is the case with ``Tap''. This overlap in tactile cues may lead to confusion and reduce the fidelity of gesture recognition when conveying gestures to participants. Overall, participants successfully decoded all the touch gestures, and touch gesture decoding performed better than emotion decoding. 

\begin{figure}
    \centering
    \includegraphics[height=4.6cm]{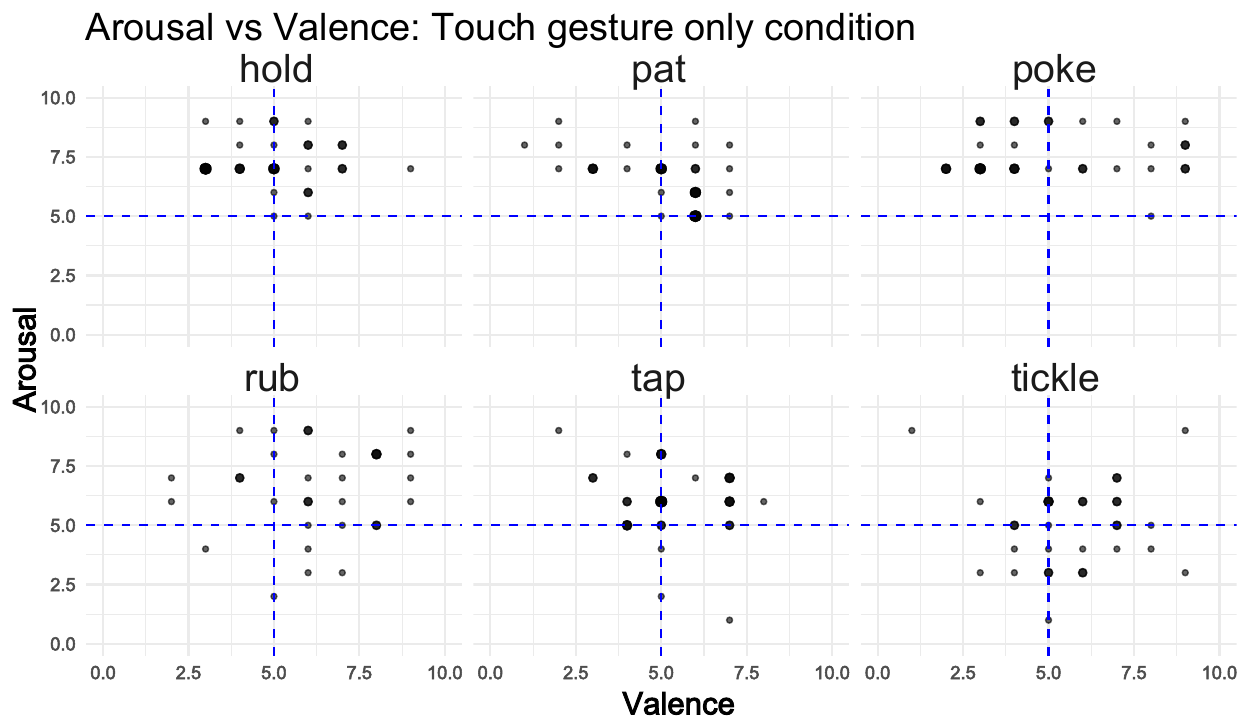}
    \caption{Scatter plots for affective response of touch gestures in the touch modality.}
    \label{fig:scatter_gesture_touch}
\end{figure}

In terms of affective responses to touch gestures in the touch modality, as shown in Fig.~\ref{fig:scatter_gesture_touch} and Table.~\ref {tab:gesture_touch}, gestures such as ``Hold'', ``Poke'', ``Rub'', and ``Tap'' were predominantly associated with high arousal and positive valence. Interestingly, ``Pat'' was classified as high arousal but neutral in valence, while ``Tickle'' contrary to expectations, was placed in the low arousal, positive valence quadrant. This classification differs notably from the sound modality, where the same gestures tended to elicit higher arousal ratings. One possible explanation for this discrepancy is that tactile perception focuses more on contact dynamics and pressure, while auditory cues (such as rhythm and intensity of impact sounds) may amplify the perceived energy or emotional expressiveness of the gesture, leading to higher arousal interpretations. Additionally, the lack of accompanying sound in the touch modality may soften the perceived intensity of certain gestures, such as ``Tickle'', making them feel more soothing than stimulating.

\subsection{Combined Sound and touch modality}

\begin{figure}
    \centering
    \centering
    \includegraphics[width=\columnwidth]{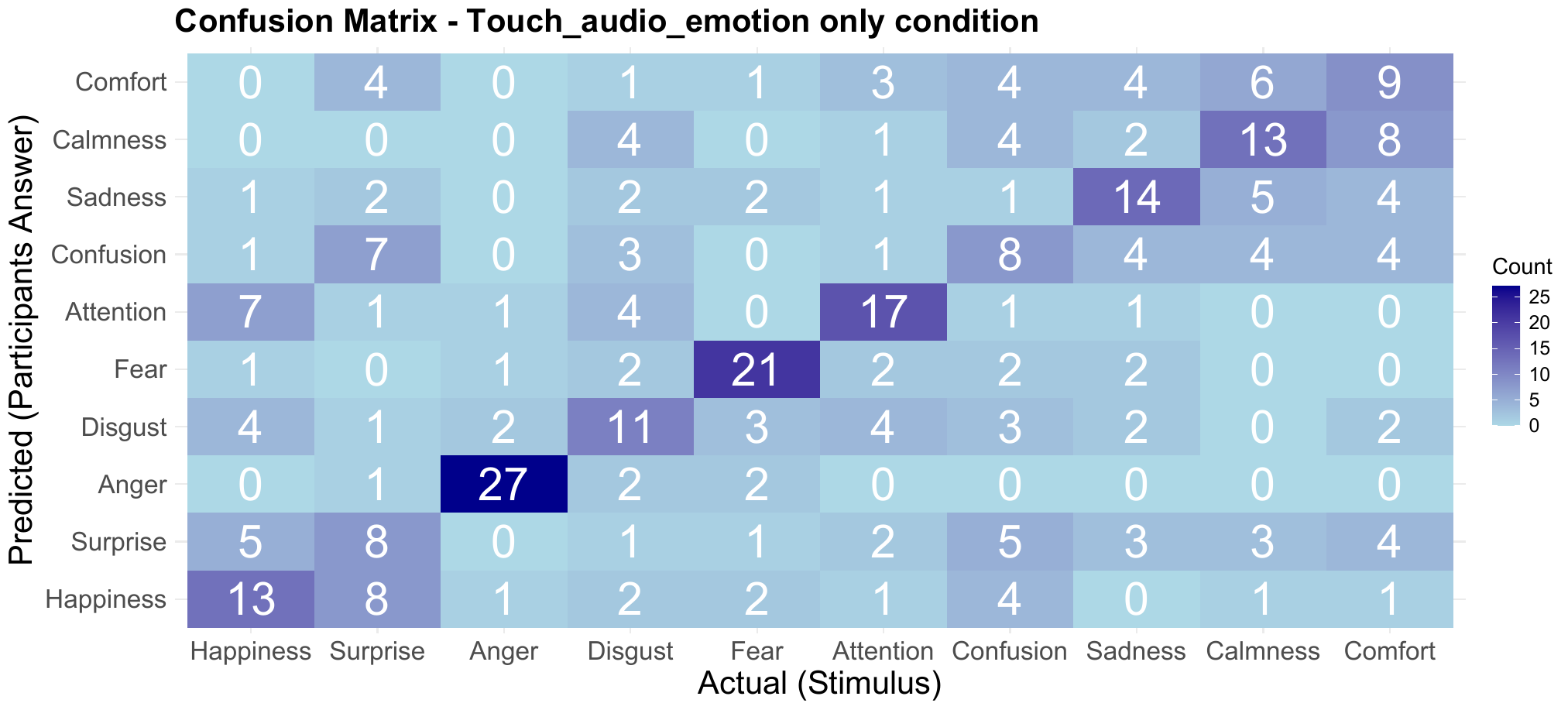}
\caption{Confusion matrix for emotions (sound and touch modality).}
    \label{fig:emotion_confusion_both}
\end{figure}

We averaged participants' arousal and valence ratings for different emotional stimuli during the \textit{Combined Touch and Sound Modality} session. The results for the 10 emotions are shown in Table.~\ref{tab:emotion_both}. 
Figure.~\ref{fig:emotion_confusion_both} illustrates the distribution of the 10 emotions within Russell's circumplex model.

\begin{table*}\footnotesize
\setlength{\abovecaptionskip}{0.0cm}   
	\setlength{\belowcaptionskip}{-0cm}  
	\renewcommand\tabcolsep{2.0pt} 
	\centering
	\caption{Arousal and valence of different emotions and decoding accuracy(DBA,\%) based on sound and touch modality. To test whether decoding accuracy surpassed chance performance (10\%), we conducted one-sample binomial tests for each emotion. Reported p-values reflect these comparisons, and significant results are marked in bold.}
	\begin{tabular}
	{
	p{1.5cm}<{\centering} 
 p{1cm}<{\centering} 
	 p{1.5cm}<{\centering} 
  p{1.5cm}<{\centering}
	p{1cm}<{\centering}
 	p{1.5cm}<{\centering}
        p{1cm}<{\centering}
 	p{1.5cm}<{\centering}
 	p{1.5cm}<{\centering}
 	p{1.5cm}<{\centering}
 	p{1cm}<{\centering}
	} 
\hline

     {Emotions} & {Happiness} & {Surprise}  & {Fear} & {Disgust} & {Anger} & {Comfort}  & {Attention} & {Calming} & {Confusion} & {Sadness} \\

\hline
   {Arousal} & {$7.2\pm1.5$} & {$5.9\pm1.9$} & {$7.0\pm2.1$} & {$5.5\pm2.0$} & {$8.8\pm0.6$} & {$4.4\pm2.4$} & {$5.5\pm1.5$} & {$4.9\pm1.8$} & {$4.8\pm1.8$} & {$4.5\pm1.8$}\\
   {Valence} & {$5.9\pm2.1$} & {$5.2\pm2.2$} & {$2.6\pm1.4$} & {$4.5\pm2.0$} & {$1.2\pm0.4$} & {$6.2\pm1.6$} & {$5.0\pm1.7$} & {$5.2\pm1.9$} & {$4.7\pm2.2$} & {$4.2\pm2.1$}\\
    {Accuracy} & {$\textbf{40.6}$} & {$\textbf{25.0}$} & {$\textbf{65.6}$} & {$\textbf{34.4}$} & {$\textbf{84.4}$} & {$\textbf{28.1}$} & {$\textbf{53.1}$} & {$\textbf{40.6}$} & {$\textbf{25.0}$} & {$\textbf{43.8}$}\\
    {Sig.(p)} & {$<0.01$} & {$0.03$} & {$<0.01$} & {$<0.01$} & {$<0.01$} & {$0.02$} & {$<0.01$} & {$<0.01$} & {$0.03$} & {$<0.01$}\\

     \hline

    \end{tabular}
\label{tab:emotion_both}
\end{table*}

Participants placed ``Happiness'' and ``Surprise'' in the high-arousal, positive-valence quadrant. ``Disgust'', ``Fear'', ``Surprise'', and ``Anger'' were categorized in the high-arousal, negative-valence quadrant. ``Confusion'' and ``Sadness'' were assigned to the low-arousal, negative-valence quadrant, while ``Comfort'' and ``Calming'' were grouped in the low-arousal, positive-valence quadrant. This classification is consistent with the emotion categorization described in Russell's circumplex model and previous research \cite{noordewier2013valence, bradley2001emotion}.

The overall decoding accuracy across the 10 emotions is 44.1\%, and the pair decoding results in Fig.~\ref{fig:emotion_confusion_both} showed that  27 out of 32 participants correctly identified ``Anger'', making it the most accurately decoded emotion. This was followed by ``Fear'', which 21 participants decoded correctly, and ``Attention'', which was decoded correctly by 17 participants. 

\begin{figure}
    \centering
    \includegraphics[width=\columnwidth]{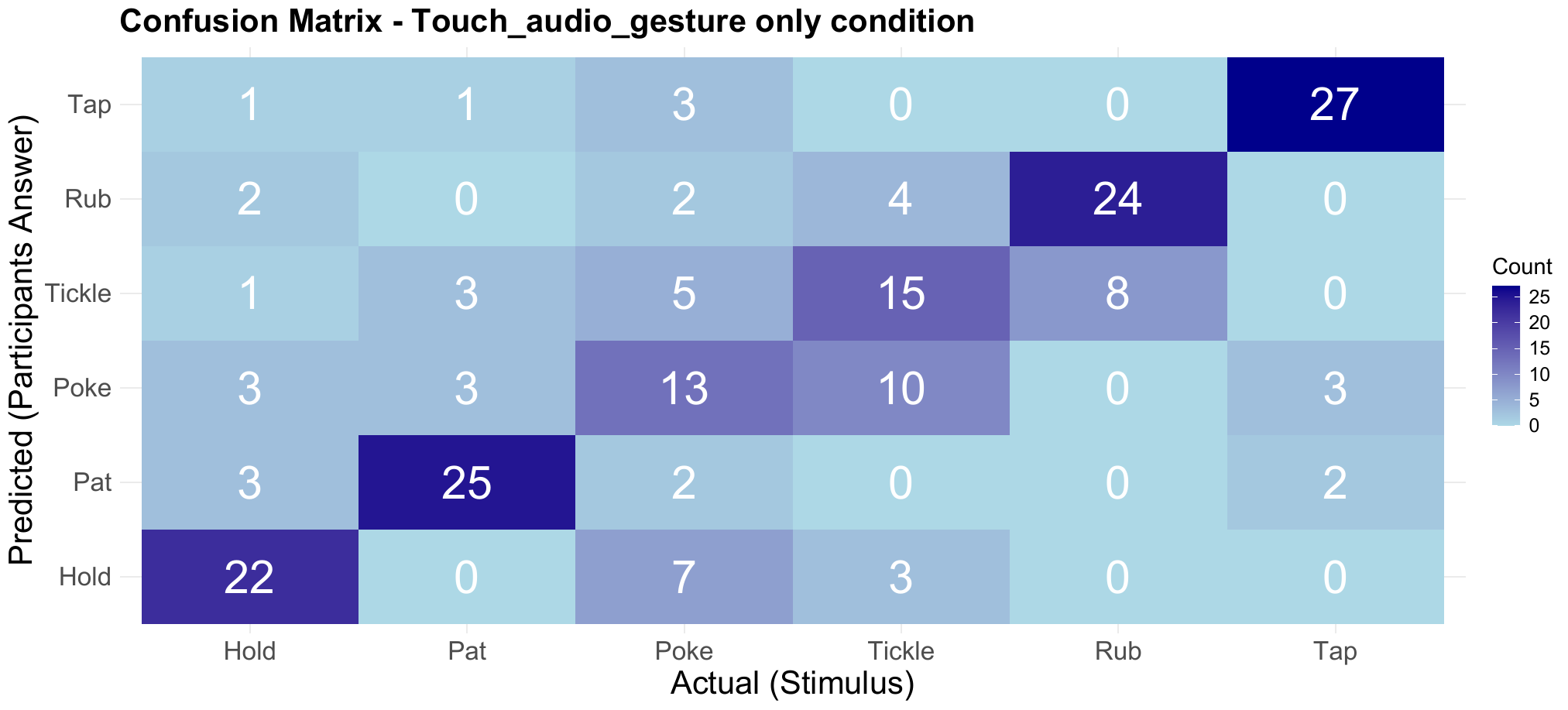}
    \caption{Confusion matrix for gestures (combined sound and touch modality).}
    \label{fig:gesture_confusion_both}
\end{figure}

As shown in Table.~\ref{tab:gesture_touch} and Fig.~\ref{fig:gesture_confusion_both}, the gesture that participants found easiest to decode was ``Tap'', followed by ``Rub'' and ``Pat''. In general, the gesture decoding accuracy is quite good, except ``Poke'' is sometimes misclassified as ``Tickle''. The participants' touch gesture decoding accuracy is 66.1\%. Overall, participants decoded all the touch gestures significantly higher than the chance level, and touch gesture decoding performed better than emotion decoding.

\begin{table}
\footnotesize
\setlength{\abovecaptionskip}{0.0cm}   
	\setlength{\belowcaptionskip}{-0cm}  
	\renewcommand\tabcolsep{2.0pt} 
	\centering
	\caption{Decoding accuracy(\%) based on sound and touch modality. To test whether decoding accuracy was above chance (16.7\%), one-sample binomial tests were performed for each gesture. Reported p-values reflect these comparisons.}
	\begin{tabular}
	{
	p{1.7cm}<{\centering} 
 p{0.9cm}<{\centering} 
	 p{1.4cm}<{\centering} 
  p{0.9cm}<{\centering}
	p{0.9cm}<{\centering}
 	p{0.9cm}<{\centering}
        p{0.9cm}<{\centering}
	} 
\hline
      
     {Gestures} & {Hold} & {Pat}  & {Poke} & {Rub} & {Tap} & {Tickle}  \\
     
\hline
   {Accuracy} & {$68.8$} & {$78.1$} & {$40.6$} & {$75.0$} & {$84.4$} & {$46.9$} \\
    {Sig.(p)} & {$<0.01$} & {$<0.01$} & {$<0.01$} & {$<0.01$} & {$<0.01$} & {$<0.01$} \\
    \hline
     {Arousal\_mean} & {$6.97$} & {$5.97$} & {$7.06$} & {$5.97$} & {$5.38$} & {$5.72$} \\
    {Arousal\_std} & {$1.56$} & {$1.75$} & {$1.52$} & {$1.84$} & {$1.66$} & {$1.55$} \\
    \hline
    {Valence\_mean} & {$4.72$} & {$5.38$} & {$4.59$} & {$5.78$} & {$5.72$} & {$5.19$} \\
    {Valence\_std} & {$1.85$} & {$1.81$} & {$1.74$} & {$1.62$} & {$1.55$} & {$2.13$} \\

     \hline

    \end{tabular}
\label{tab:gesture_both}
\end{table}

\begin{figure}
    \centering
    \includegraphics[height=4.6cm]{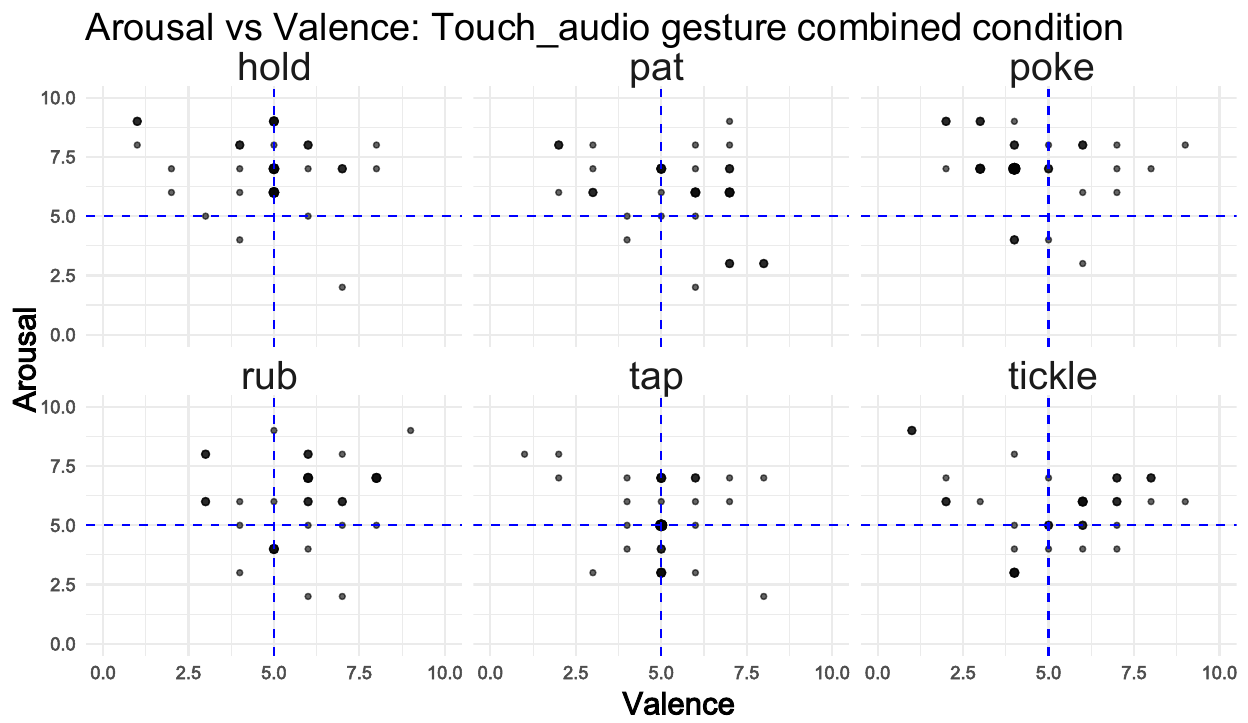}
    \caption{Scatter plots for affective response of touch gestures in the combined touch and sound modality.}
    \label{fig:scatter_gesture_both}
\end{figure}

For the multimodal condition, the affective responses to touch gestures—presented in Fig.~\ref{fig:scatter_gesture_both} and Table.~\ref{tab:gesture_both}—revealed distinct patterns across the arousal–valence space. Specifically, ``Hold'' and ``Poke'' were classified in the high arousal, negative valence quadrant, suggesting that these gestures were perceived as intense but emotionally aversive or uncomfortable when accompanied by both tactile and auditory cues. In contrast, ``Pat'', ``Rub'', ``Tap'', and ``Tickle'' were positioned in the high arousal, positive valence quadrant, indicating that these gestures were interpreted as pleasant.

This distribution highlights the modulating effect of multimodal feedback. The addition of sound appears to sharpen emotional interpretation, possibly by reinforcing or clarifying the intent behind each gesture. For example, when we try to convey ``Hold'' gesture through vibration, as this gesture trigger more vibration motors than other gesture, like ``Poke'' or ``Tap'', and the amplitude is high, people might interpret it as negative valence and high arousal; however, the generated sound might be very little, so it might cause some conflict. On the contrary, ``Poke'', typically brief and neutral, may be interpreted as more abrupt or even aggressive when its associated sound is amplified. On the other hand, rhythmic and familiar gestures like ``Pat'' or ``Tap'' benefit from congruent audio-tactile feedback, enhancing their positive affective interpretation. Overall, these findings underscore the importance of sensory synergy in shaping emotional responses and emphasize the need to carefully calibrate multimodal cues when designing robot touch behaviours that are intended to convey specific social or emotional meanings.

\subsection{Qualitative results and analysis}

\subsubsection{Questionnaire}

50\% of participants reported that ``Disgust'' was the most difficult emotion to decode, followed by ``Confusion'' (21.98\%), and ``Happiness'' (9.4\%). In contrast, none of the participants selected ``Grab attention'' or ``Anger'' as difficult to interpret, which aligns with their decoding results and suggests that participants were able to identify consistent cues associated with these two emotions.

As for emotion decoding from the touch modality, none of the participants reported being very confident. Approximately 53.1\% of participants rated their confidence as 2 out of 5, while 18.8\% indicated no confidence at all (1 out of 5). In comparison, participants expressed greater confidence in decoding emotions from the sound modality, with the average confidence rating being notably higher. Specifically, 72.7\% of participants rated their confidence as 4 out of 5.

Regarding the strategies used for emotion decoding, 40.6\% of participants reported relying on a combination of vibration features (e.g., intensity, magnitude, frequency, speed), sound characteristics (e.g., intensity or pitch), and the speed or rhythm of the gesture. Additionally, 37.5\% of participants indicated using a broader set of cues, including vibration and sound features, gesture dynamics, and the specific touch gestures they perceived.

Participants' responses revealed notable subjective confusion when decoding the emotions conveyed through touch and sound modality. Rather than offering consistent interpretations, many participants explicitly reported difficulty distinguishing between specific emotions or gestures, indicating that their perception of affective cues was often ambiguous or overlapping. Some participants mentioned being confused between emotions such as ``Comfort'', ``Sadness'', ``Calmness'', and ``Confusion''. These emotions all belong to the low-arousal emotions. One participant reported emotions like ``Anger'' and ``Disgust'' are also easily misclassified, and those two emotions are high-arousal emotions. This suggests that participants experienced difficulty in differentiating between emotional expressions that share similar valence or arousal levels, particularly those with low arousal and soft tactile features. For example, a slow, gentle vibration might be interpreted as calming by one participant and melancholic by another, depending on their personal emotional framework. Some participants explicitly noted confusions like ``Disgust'' with ``Anger'' or ``Surprise'' with ``Calming''. These confusions highlight how similar tactile characteristics across certain emotions affected participants' ability to decode them accurately. Although participants recognized these emotions as distinct, pinpointing explicit tactile differences between them remained challenging.

In addition to confusion on emotion decoding, some participants reported uncertainty between similar gestures, such as tap and pat. These gestures may share similar tactile characteristics, such as brief contact or repetitive motion, leading to inconsistent emotional interpretations. As a result, the perceived emotion may be influenced more by gesture familiarity or personal associations than by any clear expressive intent. The participant feedback underscores that emotion decoding in this context is not a purely objective or consistent process. Individual experiences, cultural backgrounds, and expectations all shape how people interpret affective touch and sound. These reports of confusion highlight that some tactile signals may lack the distinctiveness or clarity needed to reliably convey specific emotional states across different individuals.


\section{Discussion}

This study investigated how people interpret emotions and touch gestures in human–robot interaction, and whether combining auditory and haptic cues improves recognition accuracy compared to using each modality alone. Our results clearly show that participants were able to recognize both emotions and gestures using all three conditions—haptic, auditory, and their combination—though with varying levels of accuracy and modality-specific strengths.

Emotion recognition was most effective in the multimodal condition, where participants achieved an accuracy of 44.1\%, significantly outperforming the sound-only (31.6\%) and touch-only (25\%) conditions. The results highlight the complementary nature of auditory and haptic channels. Auditory cues provide high-resolution temporal and spectral features, such as pitch, loudness, and rhythm, which humans are evolutionarily tuned to interpret for emotional content \cite{bryant2013animal}. Haptic cues, in contrast, deliver spatial, force-based, and rhythmic feedback directly to the body, offering a more embodied and visceral experience of affect \cite{ren2024conveying}.

Importantly, each single modality also showed unique advantages, which can be seen also in \cite{ezzameli2023emotion}. The auditory channel enabled higher confidence ratings from participants and generally higher emotion recognition accuracy. It was particularly effective for high-arousal emotions such as Anger, Fear, and Surprise, likely due to their characteristic sound dynamics (e.g., sharp, loud, or abrupt auditory features). On the other hand, the haptic channel showed greater success in conveying affective nuances for gestures and for emotions involving bodily proximity, such as ``Attention'' or ``Fear''. While decoding accuracy was lower than with sound, haptic feedback was still effective—especially when spatial vibration patterns or rhythms were clearly defined.

Gesture recognition benefited even more strongly from multimodal input. In the combined condition, gestures such as ``Tap'', ``Rub'', and ``Hold'' were identified with accuracies up to 84.4\%, whereas even in the touch-only condition, recognition remained above chance. This suggests that motor-based pattern recognition—especially with consistent rhythm and spatial placement—is intuitively accessible through the tactile sense. However, certain emotional categories—especially those within similar valence–arousal zones (e.g., ``Calming'', ``Comfort'', ``Sadness'', ``Confusion'') — remained challenging across all modalities. These results confirm that affective touch decoding is subjective, context-sensitive, and vulnerable to overlap when relying on a single sensory channel. Multimodal feedback, by reinforcing emotional signals through redundancy and cross-modal convergence, helps disambiguate such cases.

These results reinforce a key insight: emotional communication through touch is deeply subjective, especially in the absence of contextual or visual cues. Participants frequently reported difficulty distinguishing between low-arousal emotions such as ``Comfort'', ``Calming'', ``Sadness'', and ``Confusion'', which tend to share overlapping tactile qualities (e.g., gentle pressure, slow rhythm). Such perceptual overlap highlights a fundamental challenge in affective robotics: while human touch conveys rich emotional meaning in interpersonal interaction, it is highly context-dependent, influenced by cultural background, personal associations, and situational interpretation \cite{ren2025situated}.

Our analysis of the individual contributions of haptic and auditory channels to emotion and gesture decoding, and how their effectiveness varies by expression, reveals notable differences in user experience and modality performance. Participants reported significantly higher confidence in decoding emotions via the sound modality, with 72.7\% rating their confidence at 4 out of 5. In contrast, tactile decoding was met with much more uncertainty: 53.1\% of participants rated their confidence as only 2 out of 5, and 18.8\% reported no confidence at all. This discrepancy likely reflects the human auditory system’s superior temporal and spectral resolution, which makes it easier to detect emotional nuances through pitch, volume, and rhythm. Tactile sensitivity, on the other hand, varies widely between individuals and across different body areas, making touch-based emotional interpretation more subjective and inconsistent \cite{serino2010touch}, which might extend to human-robot interaction as well.

Despite this, the haptic channel still showed unique strengths for certain emotions. Fear and Happiness, for instance, were recognized with relatively high accuracy in the touch-only condition (40.6\% and 34.4\%, respectively). Participants reported relying on tactile features such as vibration frequency, intensity, and gesture speed to make judgments, indicating that touch, when well-calibrated, can effectively convey emotional urgency or positivity. However, some high-arousal emotions like ``Anger'' and ``Disgust'' were frequently misinterpreted across all modalities. This may be due to overlapping tactile features, such as strong or fast vibrations, which fail to distinguish between different negative affective states.

As for the confusion between different emotions. Emotions like ``Disgust'', ``Confusion'', and ``Sadness'' were among the most difficult for participants to interpret. Over half reported that More than half of the participants reported that ``Disgust'' was the most difficult emotion to decode. This difficulty likely stems from the subtle or ambiguous tactile and auditory cues generated by the chosen gesture (e.g., a slow pushing-away movement). Moreover, participants held different interpretations of how ``Disgust'' should be expressed through touch. Some felt that the robot should not touch at all when conveying this emotion, while others expected a clear pushing-away gesture. Still others considered it similar to ``Anger'', anticipating strong and intense vibrations. These divergent expectations highlight a lack of consistency among participants in how they conceptualize tactile expressions of Disgust, which may explain the confusion. ``Confusion'' also frequently overlapped with ``Comfort'', ``Calming'', and ``Sadness'', highlighting how emotions within the same low-arousal, similar-valence space are particularly prone to perceptual blending.

In terms of gestures, ``Tap'', ``Pat'' and ``Poke'' were often confused, as they all involved brief, rhythmic contact with overlapping spatial patterns. The limited surface area of the vibration sleeve may have further contributed to the difficulty in distinguishing between these gestures. These findings suggest that fine distinctions in gesture dynamics are hard to perceive unless the haptic feedback system has high spatial resolution and a broader range of expression. Improving the granularity and richness of tactile feedback—such as by varying motor location, amplitude contrast, or rhythm complexity—may help reduce confusion and improve decoding accuracy across both emotional and gestural domains.

\subsection{Limitations}

While this study provides valuable insights into multimodal emotion and gesture recognition in human–robot interaction, several limitations should be acknowledged:

All participants were recruited from a single cultural background (Chinese), which may limit the generalizability of the findings. Cultural norms and expectations around touch and emotional expression vary significantly, and future studies should include more diverse populations to explore cross-cultural differences in affective interpretation.

The study focused on ten emotions and six predefined social touch gestures. While this set captures a meaningful range of affective expressions, it does not cover the full complexity of human emotional and social communication. Emotions like pride, embarrassment, or empathy, which may involve more subtle cues, were not included.

The tactile feedback was delivered using a 5×5 vibration motor sleeve with limited surface area and resolution. As participants noted, the small contact area may have made it difficult to distinguish between similar gestures (e.g., tap vs. pat), potentially lowering decoding accuracy. Additionally, the latency and limited frequency range of the vibration motors may have dampened the expressivity of certain emotional signals, such as anger or urgency.

In real-life interactions, emotional expressions are often interpreted within rich social, visual, and contextual frameworks. Our study isolated sound and touch stimuli from contextual cues to focus on modality-specific decoding, but this may have made some emotions—particularly low-arousal or ambiguous ones—more difficult to interpret. Integrating situational or embodied context in future experiments could provide a more ecologically valid understanding of multimodal affect decoding.

While participants were provided with standardized emotion and gesture definitions, their interpretations and decoding strategies were inevitably influenced by individual experiences, expectations, and preferences. As our qualitative results suggest, emotional perception is inherently subjective, and some participants relied on intuition or gut feeling rather than systematic decoding strategies.

All participants received the same stimuli derived from selected representative samples. Although this ensured consistency in presentation, it may have limited the variability and richness of expression seen in more naturalistic or spontaneous interactions. Future studies could explore participant-generated stimuli or real-time robot expression to increase ecological relevance.


During the experiment, participants rarely looked at the robot, even though it was present in the room. One possible explanation is that the study focused on the sound and touch modalities, and participants did not have direct communication with the robot. However, the influence of the robot’s appearance remains unexplored. Similarly, the potential effect of the robot’s facial expressions has not been addressed; in this experiment, we used a neutral facial expression. Since this research focused exclusively on sound and touch modalities, the interpretation of results might differ if a visual modality were also involved.



\section{Conclusion}

This study explored how humans decode emotions and social touch gestures conveyed by a robot through auditory, haptic, and combined modalities. Our results show that while each modality contributes uniquely to affective interpretation, multimodal integration significantly enhances decoding accuracy, especially for complex emotional expressions. Participants could reliably distinguish emotions across different arousal–valence zones, but decoding became more difficult when emotions shared similar affective profiles. Gesture recognition was generally more accurate than emotion recognition, particularly in the combined modality. These findings highlight the value of multisensory feedback in human–robot interaction and support the design of socially expressive robots that leverage both touch and sound for more intuitive and emotionally resonant communication.



\bibliographystyle{ieeetr}
\bibliography{references}

\vfill

\end{document}